\documentclass{article}
\PassOptionsToPackage{numbers, compress}{natbib}

    \usepackage[preprint]{neurips_2025}
\usepackage{listings}

\usepackage[utf8]{inputenc} % allow utf-8 input
\usepackage[T1]{fontenc}    % use 8-bit T1 fonts
\usepackage{hyperref}       % hyperlinks
\usepackage{url}            % simple URL typesetting
\usepackage{booktabs}       % professional-quality tables
\usepackage{amsfonts}       % blackboard math symbols
\usepackage[ruled,vlined]{algorithm2e} % for algorithm environment
\usepackage[table]{xcolor} % color for tables
\usepackage{amsthm}
\newtheorem{definition}{Definition}
\newtheorem{remark}{Remark}
\usepackage{nicefrac}       % compact symbols for 1/2, etc.
\usepackage{microtype}      % microtypography
\usepackage{xcolor}         % colors
\usepackage{graphicx}
\usepackage{amsmath}
\usepackage{amssymb}
\usepackage{pifont}
\usepackage{multirow}
\usepackage{caption}
\newcommand{\checkmarks}{\ding{51}}
\newcommand{\xmark}{\ding{55}}
\definecolor{mygray}{gray}{.9}
\newcommand{\lei}[1]{\begin{color}{black}#1\end{color}}
\usepackage{wrapfig}
\title{%
  \raisebox{-0.35\height}{\includegraphics[width=2.0em]{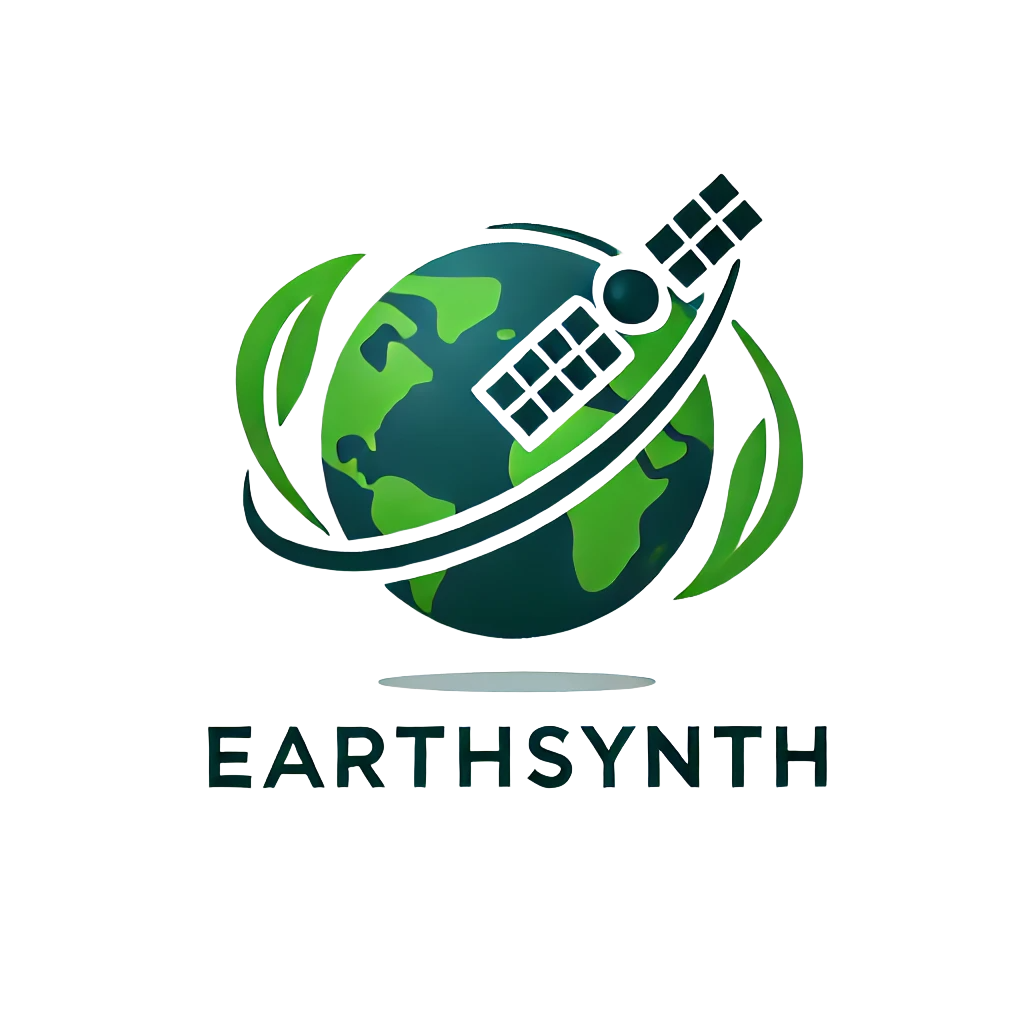}} % 图标
  EarthSynth: Generating Informative Earth Observation with Diffusion Models
}

\author{%
  Jiancheng~Pan$^{1,3,*}$,
  Shiye~Lei$^{2,}$\thanks{Contributed equally.},
  Yuqian~Fu$^{3,\dagger}$,
  Jiahao~Li$^{1,}$,
  Yanxing~Liu$^{4}$, \\
  \textbf{Yuze~Sun}$^{1}$,
  \textbf{Xiao~He}$^{5}$,
  \textbf{Long~Peng}$^{6}$,
  \textbf{Xiaomeng~Huang}$^{1,\dagger}$,
  \textbf{Bo~Zhao}$^{7,}$\thanks{Corresponding author.} \\
  $^{1}$ Tsinghua University, $^{2}$ University of Sydney, \\ $^{3}$ INSAIT, Sofia University ``St. Kliment Ohridski'' \\ $^{4}$ University of Chinese Academy of Sciences, $^{5}$ Wuhan University,\\ $^{6}$ University of Science and Technology of China, $^{7}$ Shanghai Jiao Tong University\\
  Project Page: \textcolor{magenta}{https://jaychempan.github.io/EarthSynth-website}
}

\begin{document}

\maketitle

\begin{figure}[ht]
% \vskip 0.1in
\begin{center}
\centerline{\includegraphics[width=0.8\linewidth]{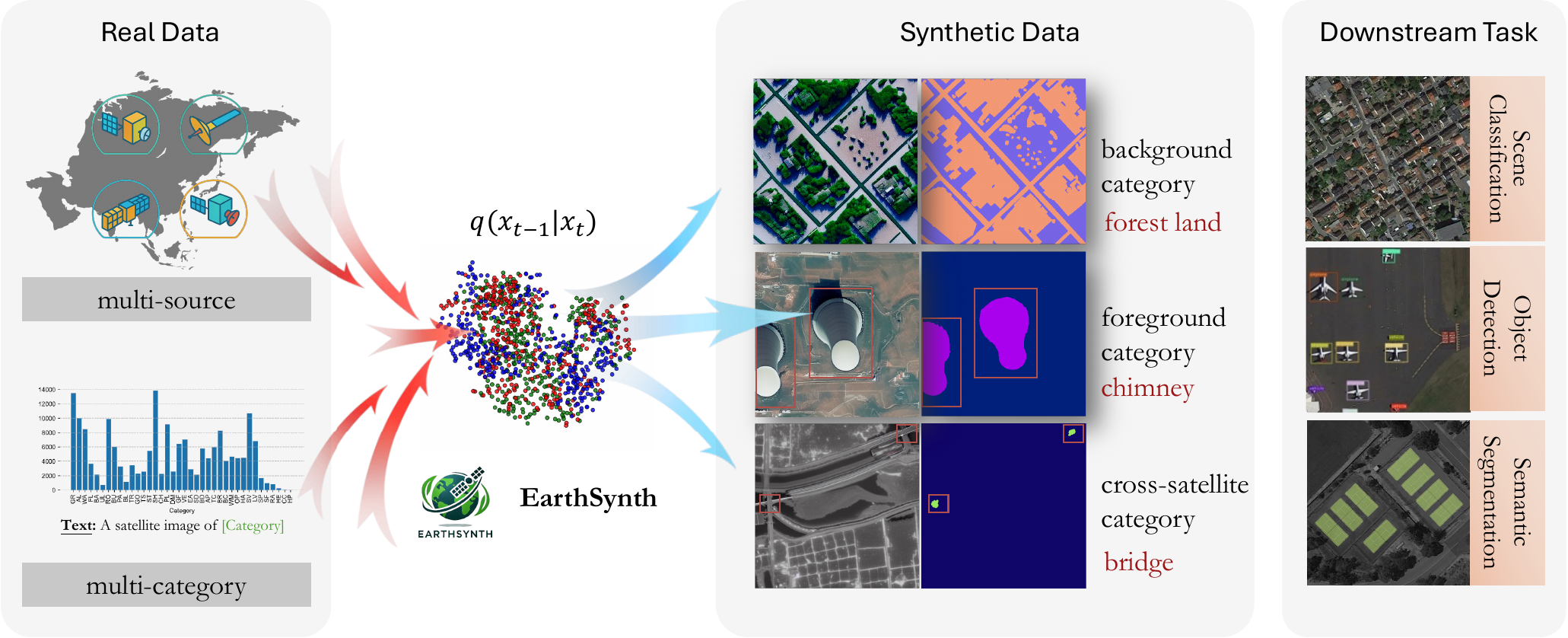}}
\caption{A diffusion-based generative foundation model, EarthSynth, pretrained on multi-source and multi-category data, synthesizing Earth observation with a semantic mask and text for downstream remote sensing image interpretation tasks.}
\label{fig:fig1}
\end{center}
% \vskip -0.1in
\end{figure}

\begin{abstract}
Remote sensing image (RSI) interpretation typically faces challenges due to the scarcity of labeled data, which limits the performance of RSI interpretation tasks. To tackle this challenge, we propose \textbf{EarthSynth}, a diffusion-based generative foundation model that enables synthesizing multi-category, cross-satellite labeled Earth observation for downstream RSI interpretation tasks. To the best of our knowledge, EarthSynth is the \textit{first to explore multi-task generation for remote sensing}, tackling the challenge of limited generalization in task-oriented synthesis for RSI interpretation. EarthSynth, trained on the EarthSynth-180K dataset, employs the Counterfactual Composition training strategy with a three-dimensional batch-sample selection mechanism to improve training data diversity and enhance category control. Furthermore, a rule-based method of R-Filter is proposed to filter more informative synthetic data for downstream tasks. We evaluate our EarthSynth on scene classification, object detection, and semantic segmentation in open-world scenarios. There are significant improvements in open-vocabulary understanding tasks, offering a practical solution for advancing RSI interpretation.
\end{abstract}

\section{Introduction}\label{sec:sec1}
\textcolor{black}{Remote sensing image (RSI) interpretation~\cite{sun2021research} is fundamentally constrained by challenges such as severe class imbalance~\cite{sharma2025addressing} and limited availability of high-quality labeled data, significantly impeding the development of robust models for downstream tasks. However, labeling RSIs typically requires domain-specific expertise and substantial manual effort, making large-scale annotation time-consuming and costly. Consequently, an important research objective is to effectively exploit existing labeled Earth observation (EO) datasets by uncovering latent relationships among samples to improve data efficiency.}

\textcolor{black}{In parallel, recent advances in generative data augmentation~\cite{antoniou2017data}, particularly Diffusion Models (DMs)~\cite{rombach2022high}, offer a promising avenue for synthesizing high-quality labeled training data. Data augmentation using DMs has been widely applied across various domains~\cite{chen2024comprehensive}. Prior works~\cite{he2022synthetic,trabucco2023effective} investigate text-to-image (T2I) models~\cite{huang2024learning} to boost classification performance in data-scarce scenarios, enlightening and inspiring the remote sensing community. And some efforts~\cite{alemohammad2024self,alemohammad2024selfconsuming} research the risk of model collapse and improve generation quality through self-generated data. By augmenting rare classes and enriching data diversity~\cite{NEURIPS2023_f99f7b22}, these models can play a critical role in mitigating data scarcity and enhancing the generalization capabilities of RSI interpretation.} Unlike existing studies, \textit{\textcolor{black}{our work centers on improving the generative diversity of diffusion models tailored for remote sensing applications.}}

\textcolor{black}{In the remote sensing community, most existing data augmentation approaches are trained predominantly on single-source EO, which inherently limits the diversity of generated categories and lacks generality and flexibility across broader remote sensing applications. However, training generative models on multi-source data requires balancing the quality and diversity of generated data. Txt2Img-MHN~\cite{xu2023txt2img} attempts to use GANs~\cite{salimans2016improved} to generate satellite images. DiffusionSat~\cite{khanna2023diffusionsat} and CRS-Diff~\cite{tang2024crs} propose conditional DMs for generating optical RSIs, incorporating diverse texture-based conditions and applying the synthesized data to downstream tasks such as road extraction. GeoSynth~\cite{toker2024satsynth} joint data manifold of images and labels for satellite semantic segmentation. AeroGen~\cite{tang2024aerogen} and MMO-IG~\cite{yang2025mmo} try to use synthesized training data for satellite object detection. However, these methods are typically trained on single tasks or single-source data, requiring repeated training and generation to meet diverse needs, which hinders their application in real-world scenarios.} 

\textcolor{black}{To solve the above problems, we propose \textbf{EarthSynth}, a diffusion-based generative foundation model, synthesizing EO with a semantic mask and text for downstream RSI interpretation tasks as shown in Figure~\ref{fig:fig1}. First, we construct the EarthSynth-180K with multi-source and
multi-category data, 180K samples for training EarthSynth. Specifically, we collect samples from multiple datasets and apply random cropping and category-augmentation strategy to standardize image resolution, ensuring alignment among images, semantic masks, and text descriptions. To our knowledge, EarthSynth-180K is the first large-scale remote sensing dataset for diffusion training. During training, we adopt the Counterfactual Composition (CF-Comp) strategy \textcolor{black}{with channel, pixel, and semantic spaces as batch-sample selection mechanism} to simultaneously enhance layout controllability and category diversity, thereby enabling the generation of more informative EO data. \textcolor{black}{Different previous studies, we apply the CF-Comp strategy to diffusion-based generative foundation model training for downtown tasks, avoiding repeating the training task-specific generative model for each downstream task.} For training data synthesis, a rule-based method of R-Filter is proposed to filter more informative synthetic data. We evaluate our EarthSynth with multiple datasets on scene classification, object detection, and semantic segmentation. Furthermore, the effectiveness of our method is demonstrated through comprehensive ablation studies and visualization analysis.}

We summarize the main contributions as follows:
\begin{itemize}
\item \textcolor{black}{We propose EarthSynth, a diffusion-based generative foundation model trained on the EarthSynth-180K dataset with 180K \textcolor{black}{cross-satellite and multi-sensor} samples aligned image, semantic mask, and text, achieving a unified solution to achieve multi-task generation.}
\item \textcolor{black}{EarthSynth employs the CF-Comp strategy to balance the layout
controllability and category diversity during training, enabling fine layout control for RSI generation. And integrates the R-Filter post-processing method to extract more informative synthesized data. }
\item \textcolor{black}{EarthSynth is evaluated on scene classification, object detection, and semantic segmentation in open-vocabulary scenarios, well validating its effectiveness.}
\end{itemize}

\section{Related Work}
\textbf{Diffusion Models for Remote Sensing.} 
Diffusion Models (DMs), which have shown great success in natural image synthesis~\cite{rombach2022high}, are increasingly being applied to remote sensing~\cite{khanna2023diffusionsat,liu2025text2earth}. Depending on the use case~\cite{liu2024diffusion}, their applications in remote sensing can be broadly divided into three categories: image enhancement, image interpretation, and image synthesis. For image enhancement, DMs improve multispectral and hyperspectral images by fusing information across channels and expanding attention~\cite{xu2023dual,feng2024multiscale,wei2024diffusion,li2023hyperspectral}. Some works also apply DMs to tasks like change detection~\cite{bandara2022ddpm} and climate prediction~\cite{gao2023prediff}, enabling better integration of multimodal data and improving pixel-level discrimination. On the other hand, image synthesis methods focus on generating artificial data to address the issue of data scarcity. Recent studies~\cite{tang2024crs,toker2024satsynth,tang2024aerogen,yang2025mmo,sastry2024geosynth} demonstrate how synthetic data can benefit downstream remote sensing tasks. Unlike prior approaches that fine-tune on a single-source dataset for one specific task, we adopt a more general training and synthesis strategy. Notably, these data
augmentation methods~\cite{zhang2024advancing,zhang2025controllable,liu2025control,li2025domain} demonstrate significant advantages in few-shot learning tasks~\cite{fu2023styleadv,fu2025ntire}. Our approach supports multiple tasks, including scene classification, object detection, and semantic segmentation, allowing the model to generate more diverse and widely applicable data.

\begin{figure}[t]
\vskip 0.2in
\begin{center}
\centerline{\includegraphics[width=\columnwidth]{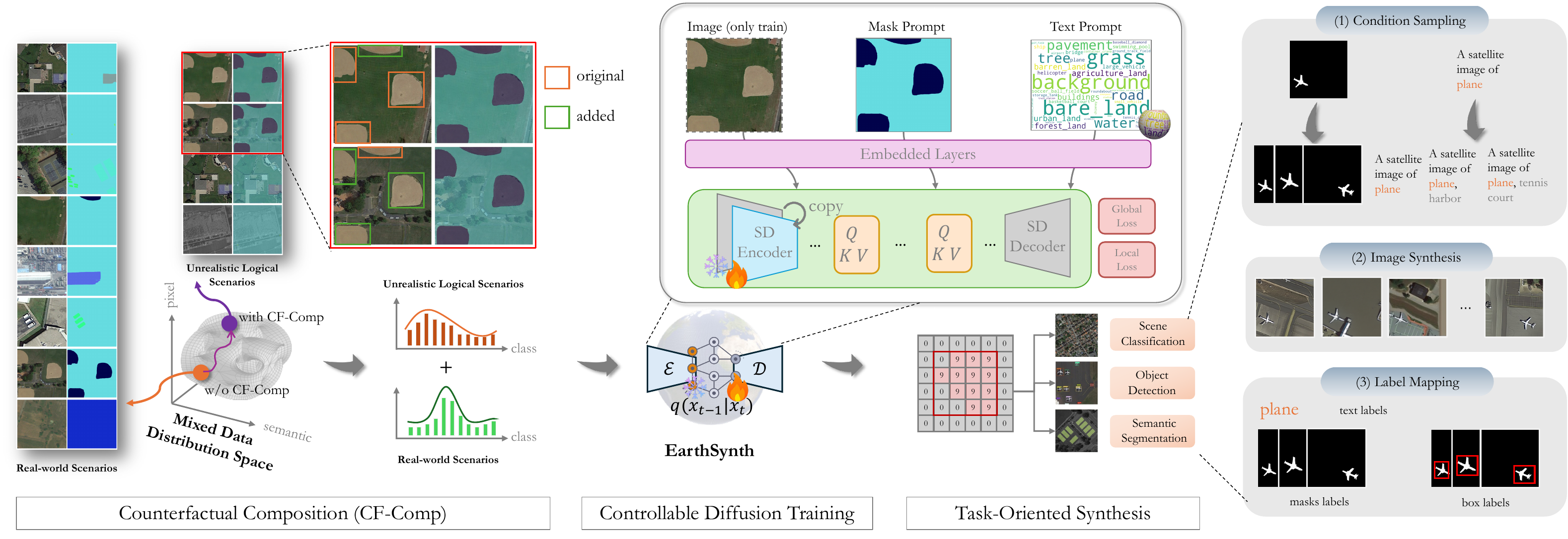}}
\caption{EarthSynth is trained with CF-Comp training strategy on real and unrealistic data distribution, learns remote sensing pixel-level properties in multiple dimensions, and builds a unified process for conditional diffusion training and synthesis.}
\label{fig:fig2}
\end{center}
\vskip -0.2in
\end{figure}

\textbf{Generative Data Augmentation with Diffusion Models.} Generative data augmentation using DMs has been explored across various domains, including natural images and remote sensing. Studies such as~\cite{he2022synthetic,trabucco2023effective} investigate using text-to-image DMs to boost classification performance under limited data conditions. To further improve generation quality,~\cite{alemohammad2024self,alemohammad2024selfconsuming} propose leveraging self-generated data; however, this approach risks model collapse due to overfitting. And some efforts~\cite{alemohammad2024self,alemohammad2024selfconsuming} research the risk of model collapse and improve generation quality through self-generated data. In remote sensing, recent works~\cite{tang2024crs,toker2024satsynth,tang2024aerogen,yang2025mmo,sastry2024geosynth, li2025domain} have begun to explore the potential of diffusion-based data synthesis and enhancement to improve RSI interpretation tasks. But there is no unified solution to achieve multi-task generation.

\section{Preliminaries}
\textbf{Training Data Synthesis.} Given a conditional generative diffusion model \( G_\theta \) with pretrained parameters $\theta$, \lei{and $x$ denotes the generative image. The the generative distribution w.r.t. $x$ induced by the condition set $\mathcal{C}$ is defined as:
\begin{equation}
\mathcal{X}_{\mathcal{C}} = \sum_{i=1}^{|\mathcal{C}|} \frac{1}{|\mathcal{C}|} \, \mathbb{P}_{G_\theta}(x \mid c_i),
\end{equation}
where $\mathcal{C} = \{c_i\}_{i=1}^{|\mathcal{C}|}$ consists of a conditional \texttt{Mask-Text} pairs $c_i$,
}
\lei{and} each conditional distribution \( \mathbb{P}_{G_\theta}(x \mid c_i) \) \textcolor{black}{is defined by generating samples $x=G_\theta(\epsilon \mid c_i)$, where \( \epsilon \sim \mathcal{N}(0, I) \)}.  The pixel-level, region-level, and semantic-level label $y=\{y_l, y_c\}=\mathcal{F}(c_i)$ is generated via a mapping function $\mathcal{F}$, where $y$ consists of a location label $y_l$ (semantic mask or bounding box) and a category label $y_c$.

\textbf{Feature Decomposition.} Feature decomposition~\cite{gao2023out} for satellite imagery $x=f\left(x_{\texttt {obj}}, x_{\texttt {bg}}, x_{\texttt {noise}}\right)$ across various categories from different satellites. In remote sensing, we can formalize some criteria by (in)dependence with label $y$ in the meta distribution $\mathbb{P}_{G_\theta}$:
\begin{equation}
x_{\texttt{obj}},x_{\texttt{bg}} \not\!\perp\!\!\!\perp y, \quad
x_{\texttt{noise}} \perp\!\!\!\perp y, \\
\end{equation}
where $y$ depends on object $x_{\texttt{obj}}$ and background $x_{\texttt{bg}}$ but is independent of noise $x_{\texttt{noise}}$ which yields $\mathbb{P}_{G_\theta}(y \mid x)=\mathbb{P}_{G_\theta}(x_{\texttt{obj}},x_{\texttt{bg}})$. $x_{\texttt{noise}}$ is the noise disturbance during satellite imaging that makes semantic confusion~\cite{pan2023reducing,pan2023prior,pan2024pir}, leading to inter-class similarity and significant intra-class variation of satellite images.

\textbf{Copy-Paste Augmentation.} Copy-Paste~\cite{ghiasi2021simple} is a data augmentation technique that involves copying objects or regions from one image and pasting them into another to create new composite scenes. As shown in Figure~\ref{fig:fig3},   $\texttt{Copy-Paste}(x^a_{obj},x^b)$ represents the operation of copying the objects of image $x^a$ to image $x^b$. \textcolor{black}{However, the Copy-Paste introduces compositional artifacts or non-smooth transitions, etc., that alter the statistical properties of the image distribution. These artifacts and transitions can typically be mitigated through the training process of DMs.} \textcolor{black}{More details are described in the Appendix.}

\section{Counterfactual Composition for Controllable Diffusion Training}
Due to the inherent characteristics of satellite imagery, satellite images often exhibit high inter-class similarity and significant intra-class variation, posing challenges for RSI interpretation. To get more informative data distribution from a generative DM, we aim to approximate a real-world distribution $\mathcal{D}^{\texttt {real}}$ with as much training data distribution $\mathcal{D}^{\texttt {train}} \subset \mathcal{D}^{\texttt {real}}$ as possible. 
Constructing training data with diverse backgrounds and objects enables the DM to learn rich and diverse semantic information in open remote sensing scenarios. To achieve this, we enhance scene diversity through counterfactual composition, which involves combining existing object categories with diverse background contexts. The definition is as follows.

\begin{figure}[t]
\vskip 0.2in
\begin{center}
\centerline{\includegraphics[width=0.9\textwidth]{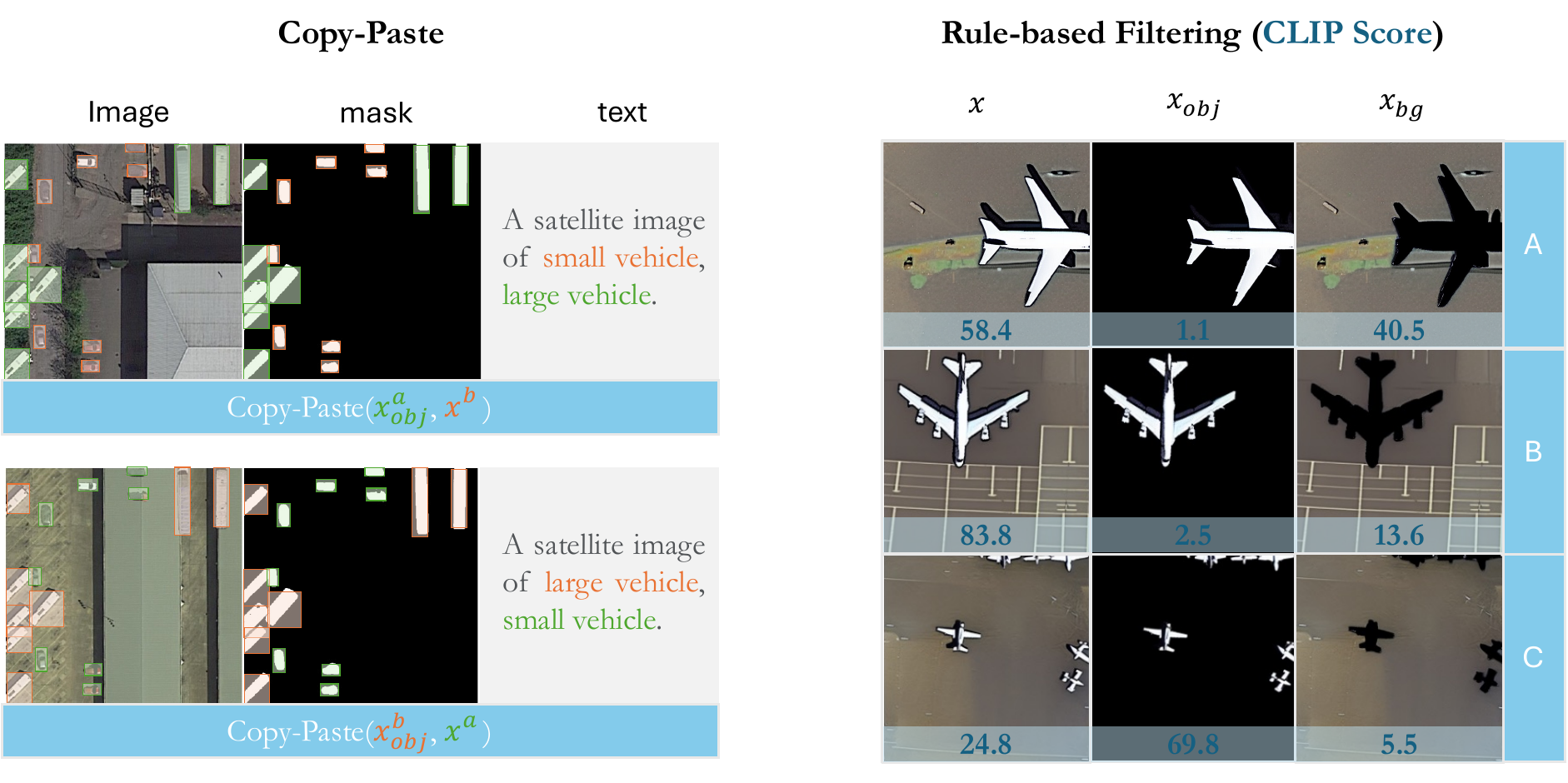}}
\caption{Left: Copy-Paste used in CF-Comp Strategy. Right: CLIP-based rule filtering retains high-quality images.}
\label{fig:fig3}
\end{center}
\vskip -0.2in
\end{figure}

\begin{definition}[Counterfactual Composition]
Given a set of source elements ${A_1, A_2, \dots, A_n}$, where each $A_i$ represents a specific semantic component (e.g., object, region, or attribute) extracted from a distinct instance, a counterfactual sample is constructed by recombining these components as:
\begin{equation}
x' = \mathcal{CF}(A_1^{(i)}, A_2^{(j)}, \dots, A_n^{(k)}). \quad i \ne j \ne k
\end{equation}
Here, $A_i^{(i)}$ denotes the $i$-th component drawn from the $i$-th source instance, and $\mathcal{CF}(\cdot)$ is a composition function that logically combines elements from different instances to form a counterfactual sample $x'$, where $x'$ is out of the distribution of $\mathcal{D}^{\texttt {real}}$, being plausible yet not observed in reality.
\end{definition}

\begin{remark}
    Counterfactual Composition allows for the generation of novel and logically consistent scenes by fusing diverse elements. The goal is to preserve semantic and structural coherence while expanding the diversity and improving the generalization of possible inputs. 
\end{remark}

\begin{algorithm}[t]
\caption{Counterfactual Composition for Controllable Diffusion Training}
\label{alg:alg1}
\KwIn{Training dataset $\mathcal{D}_{\texttt{train}} = \{(\mathcal{I}, \mathcal{M}, \mathcal{T})\}_{i=1}^{|\mathcal{D}_{\texttt{train}}|}$, where $\mathcal{I}$ is the ground-truth image, $\mathcal{M}$ is the mask prompt, and $\mathcal{T}$ is the text prompt}
\KwOut{Conditional diffusion model $\mathcal{M}_{\theta}$ with parameters $\theta$}

\For{each training step}{
    sample a mini-batch \texttt{Image-Mask-Text} triples $\mathcal{B}_{\texttt{ori}} = (\mathcal{I}, \mathcal{M}, \mathcal{T})$ from dataset $\mathcal{D}$\;

    $\triangleright$ Counterfactual Composition
    
    $B_{\texttt{copy}} \gets  \{ \}$, a thresholds $\alpha_0$, $\beta_0$, $\eta_0$
    
    \For{$(x^a, m^a, t^a), (x^b, m^b, t^b)$  in $(\mathcal{I}, \mathcal{M}, \mathcal{T}) \times (\mathcal{I}, \mathcal{M}, \mathcal{T})$ }{
    % $\triangleright$ Calculate the image color sensitivity $\alpha$ of $x^a$ and $x^b$ from the channel space dimension
    $\alpha = \texttt{ICS}(x^a, x^b)$,
    % $\triangleright$ Calculate the mask overlap rate $\beta$ of $m^a$ and $m^b$ from the pixel space dimension
    $\beta = \texttt{MOR}(m^a, m^b)$,
    % $\triangleright$ Calculate the text semantic similarity $\eta$ of $t^a$ and $t^b$ from the semantic space dimension
    $\eta = \texttt{TSS}(t^a, t^b)$;
    
    \If{$\alpha$, $\beta$, $\eta$ > $\alpha_0$, $\beta_0$, $\eta_0$}{
    
    $(x', m', t') = \texttt{Copy-Paste}(x^a_{\texttt{obj}}, x^b)$\;
    
    $B_{\texttt{copy}} \gets (x', m', t')$;
    }
    
    }
    Get a new mini-batch \texttt{Image-Mask-Text} triples $\mathcal{B} = \mathcal{B}_{\texttt{ori}} + \mathcal{B}_{\texttt{copy}}$\;

    Encode text prompt $\mathcal{T}$ using a frozen text encoder to obtain $E_\mathcal{T}$\;
    
    Sample noise $\epsilon \sim \mathcal{N}(0, 1)$ and timestep $t$\;

    Generate noisy image: $x_t = \sqrt{\alpha_t} \cdot \mathcal{I} + \sqrt{1 - \alpha_t} \cdot \epsilon$\;

    Extract mask features $F_\mathcal{M}$ from $\mathcal{M}$ using $\mathcal{M}_{\theta}$ with parameters $\theta$\;

    Inject mask features into a frozen UNet and predict noise $\hat{\epsilon}$: \\
    \hspace{1em} $\hat{\epsilon} = \text{UNet}(x_t, t, E_\mathcal{T}, \mathcal{M}_{\theta}(\mathcal{M}))$\;

    Compute global loss and local loss: $\mathcal{L} = \|\hat{\epsilon} - \epsilon\|^2 + \gamma\|F_\mathcal{M} \cdot \hat{\epsilon} - F_\mathcal{M} \cdot \epsilon\|^2$\;

    Update $\theta$ using gradient descent to minimize $\mathcal{L}$\;
}
\end{algorithm}

\textbf{Unrealistic Logical Scene.}  We aim to construct an unrealistic logical scene by counterfactual composition, i.e., satellite imagery of real-world data that does not exist in the real world but is logically present. \textcolor{black}{This is also intended to maintain consistency in the image-label union space and prevent disruptions caused by arbitrary counterfactual composition.} As shown in Figure~\ref{fig:fig2}, we use the Copy-Paste to perform counterfactual composition. For example, the unrealistic logical scene is obtained by combining the two images' small and large vehicle objects in Figure~\ref{fig:fig3}. An unrealistic logical scene can be judged from three dimensions: channel, pixel, and semantic space. We define three criteria to measure whether two images can be combined into an unrealistic logical scene:

From the channel space dimension, the Image Color Sensitivity (ICS) can be obtained as
\begin{equation}
\texttt{ICS}(x_a, x_b) = 
\begin{cases}
\mathbf{1}_{\left\{ \left| S(x^a) - S(x^b) \right| < s_0 \right\}}, & \text{if } C_a = C_b \\
0, & \text{if } C_a \ne C_b
\end{cases},
\end{equation}
where the color sensitivity $S = \operatorname{Var}(R - G) + \operatorname{Var}(R - B) + \operatorname{Var}(G - B)$, $C_a$ and $C_b$ are the number of channels of satellite images $x_a$ and $x_b$ respectively. 

From the pixel space dimension, the Mask Overlap Rate (MOR) can be obtained as
\begin{equation}
\texttt{MOR}(m^a, m^b) = 
\begin{cases}
\dfrac{|m^a \cap m^b|}{|m^a \cup m^b|}, & \text{if } |m^a \cup m^b| > 0 \\
0, & \text{otherwise}
\end{cases}.
\end{equation}

From the semantic space dimension, the text semantic similarity (TSS) can be obtained as
\begin{equation}
\texttt{TSS}(t^a, t^b) = \frac{t^a \cdot {t^b}^\top}{\|t^a\| \cdot \|t^b\|}.
\end{equation}

\begin{algorithm}[t]
\caption{Training Data Synthesis}
\label{alg:alg2}
\KwIn{Pre-trained conditional diffusion model $\mathcal{M}_{\theta}$ parameters $\theta$, sampling steps $T$, sampling condition set $\mathcal{C} = \{c_i\}_{i=1}^{|\mathcal{C}|}$}
\KwOut{Synthetic dataset $\mathcal{S} = \{x_0^{(i)}, c_i\}_{i=1}^{|\mathcal{C}|}$}

Initialize empty synthetic dataset: $\mathcal{S} \leftarrow \{ \}$\;

\For{each condition $c_i \in \mathcal{C}$}{
    Sample initial noise $x_T^{(i)} \sim \mathcal{N}(0, \mathbf{I})$\;
    
    % Encode condition $c_i$ to obtain conditioning input (e.g., text embedding $E_\mathcal{T}$, mask features)\;

    Encode text prompt $\mathcal{T}$ using a frozen text encoder to obtain $E_\mathcal{T}$\;

    Extract mask features $F_\mathcal{M}$ from $\mathcal{M}$ using $\mathcal{M}_{\theta}$ with parameters $\theta$\;
    
    \For{$t = T$ down to $1$}{
        Predict noise: $\hat{\epsilon}_t = \text{UNet}(x_t^{(i)}, t, E_\mathcal{T}, \mathcal{M}_{\theta} (c_i))$\;
        
        Update latent variable using reverse diffusion equation:\\
        \hspace{1em} $x_{t-1}^{(i)} = \frac{1}{\sqrt{\alpha_t}} \left( x_t^{(i)} - \sqrt{1 - \alpha_t} \cdot \hat{\epsilon}_t \right) + \sigma_t z$,\\
        \hspace{1em} where $z \sim \mathcal{N}(0, \mathbf{I})$ if $t > 1$, else $z = 0$\;
    }
    $\triangleright$ Rule-based Filtering
    
    \If{$Score_{\texttt{CLIP}}(x) > S_0 \mid Score_{\texttt{CLIP}}(x_{obj}) > S_0$}{
    
        Add generated sample to dataset: $\mathcal{S} \leftarrow \mathcal{S} \cup \{(x_0^{(i)}, c_i)\}$\;
    }

}

\Return Final dataset $\mathcal{S}$ sampled from conditional distribution $\mathcal{X}_{\mathcal{C}}$
\end{algorithm}

\textbf{Mixed Data Distribution.} In real remote sensing scenarios, the object and background of an image typically follow a consistent data distribution. However, in unrealistic logical data distributions, discrepancies between foreground and background distributions often lead to out-of-distribution~\cite{hsu2020generalized}. By integrating real-world and unrealistic logical data distributions, the latent data manifold becomes more complex and diverse, allowing the model to generalize better to unseen scenarios while preserving the essential characteristics of the original distribution. Assuming that the object and background of the image can be modeled as Gaussian distributions, respectively, as:

\begin{equation}
x_{\texttt{obj}} \sim \mathcal{N}(\mu_{\texttt{obj}}, \sigma^2_{\texttt{obj}} I), \quad x_{\texttt{bg}}  \sim \mathcal{N}(\mu_{\texttt{bg}},  \sigma^2_{\texttt{bg}} I),
\end{equation}
where $(\mu_{\texttt{obj}}, \sigma^2_{\texttt{obj}})$ and $(\mu_{\texttt{bg}}, \sigma^2_{\texttt{bg}})$ denote the respective mean and variance of the object and background distributions. Assuming independence and linearity of expectation, the mean and variance of the resulting image $x'$ with mask $m$ by Copy-Paste can be derived as:
\begin{equation}
\begin{aligned}
\mu_{\texttt{mix}} &= \alpha \mu_{\texttt{obj}} + (1 - \alpha) \mu_{\texttt{bg}} \\
\sigma^2_{\texttt{mix}} &= \alpha \sigma^2_{\texttt{obj}} + (1 - \alpha) \sigma^2_{\texttt{bg}} + \alpha (1 - \alpha)(\mu_{\texttt{obj}} - \mu_{\texttt{bg}})^2,
\end{aligned}
\end{equation}
where $\alpha = \frac{1}{HW} \sum_{i,j} m_{i,j}$ represents the proportion of foreground pixels in the image. The additional term $\alpha(1 - \alpha)(\mu_{\texttt{obj}} - \mu_{\texttt{bg}})^2$ in the variance reflects the distributional shift caused by the mismatch between object and background statistics. CF-Comp allows models to learn richer and more informative representations by controlling the Copy-Paste ratio $\Sigma \alpha$ in remote sensing. \textcolor{black}{We note that in satellite images for background distraction and intra-class variability~\cite{ma2024direction}, $\mu_{\texttt{obj}} \approx \mu_{\texttt{bg}}$ and $\sigma^2_{\texttt{obj}} \approx \sigma^2_{\texttt{bg}}$, but in natural scene images, these parameters may differ significantly, resulting in distributional shifts~\cite{huang2021importance}.}

\textbf{Global and Local Loss.} Unlike binary masks, semantic masks embed category-specific information, making pixel-level precision essential. However, using a global constraint alone is inadequate for modeling such fine-grained control. Therefore, we integrate global and local constraints to achieve a more accurate generation. We extract semantic mask features $F_\mathcal{M}$ from mask $\mathcal{M}$ using condition model $\mathcal{M}_{\theta}$ with parameters $\theta$ and inject semantic mask features into a frozen UNet~\cite{ronneberger2015u} and predict noise $\hat{\epsilon}$:
\begin{equation}
\mathcal{L_{\texttt{global}}} = \|\hat{\epsilon} - \epsilon\|^2, \quad
\mathcal{L_{\texttt{local}}} = \|F_\mathcal{M} \cdot \hat{\epsilon} - F_\mathcal{M} \cdot \epsilon\|^2, \quad
\mathcal{L} = \mathcal{L_{\texttt{global}}} + \gamma \mathcal{L_{\texttt{local}}}
\end{equation}
where $\gamma$ is the local constraint factor.

\textbf{Training Process.} Algorithm \ref{alg:alg1} shows the batch-based CF-Comp method for condition diffusion EarthSynth training. We follow Zhang et al.~\cite{zhang2023adding} to train a conditional diffusion model.

\begin{table*}[t]
\centering
\resizebox{0.9\linewidth}{!}{
\begin{tabular}{l c c c}
\toprule
\multirow{2}{*}{\textbf{Method}}  & \multicolumn{2}{c}{\textbf{Scene Classification}} \\
\cmidrule(lr){2-3} &  RSICD* (Top1 / Top5) & DIOR (Top1 / Top5) \\
\midrule
Txt2Img-MHN(VQVAE)\dag~\cite{xu2023txt2img} & 32.7 / 75.5 & - / - \\
Txt2Img-MHN(VQGAN)\dag~\cite{xu2023txt2img} & 40.9 / 72.7 & - / - \\
CRS-Diff\dag~\cite{tang2024crs} & 57.1 / 79.0 & - / - \\
StableDiffusion\dag~\cite{rombach2022high} & \textbf{61.3} / 88.3 & 41.5 / 73.0 \\
InstanceDiffusion\dag~\cite{wang2024instancediffusion}   & 59.1 / 88.1 & 44.5 / 79.0 \\
ControlNet\dag~\cite{zhang2023adding} & 55.5 / 85.5 & 46.5 / 78.5 \\
\rowcolor{blue!10}\textbf{EarthSynth (Ours)}\dag & 60.0 / \textbf{91.8} & \textbf{49.0} / \textbf{80.0} \\
\bottomrule
\end{tabular}}
\caption{\textcolor{black}{CLIP-based scene classification} accuracy on RSICD and DIOR datasets with Acc. \textcolor{black}{\dag: training on remote sensing data.}}
\label{tab:rsicd_comparison_}
\end{table*}

\begin{table*}[t]
\centering
\resizebox{0.75\linewidth}{!}{
\begin{tabular}{l l c c}
\toprule
\multirow{2}{*}{\textbf{Method}} & \multirow{2}{*}{\textbf{Data Usage}} & \multicolumn{2}{c}{\textbf{Object Detection}} \\
\cmidrule(lr){3-4}
&  & DOTAv2 & DIOR\\
\midrule
Base GroundingDINO & \textit{Real} & 56.3 & 74.0 \\
+ StableDiffusion~\cite{rombach2022high} & \textit{Real} + \textit{Synth} & - & - \\
+ ControlNet~\cite{zhang2023adding} & \textit{Real} + \textit{Synth} & 57.4 (\textcolor{blue}{+1.1}) & 74.1 (\textcolor{blue}{+0.1}) \\
% \rowcolor{gray!20}EarthSynth + GroundingDINO & \textit{Real} + \textit{Synth} & \textbf{58.4} (\textcolor{blue}{+2.1}) & \textbf{74.3} (\textcolor{blue}{+0.3}) \\
\rowcolor{blue!10} + \textbf{EarthSynth (Ours)} & \textit{Real} + \textit{Synth} & \textbf{58.4} (\textcolor{blue}{+2.1}) & \textbf{74.3} (\textcolor{blue}{+0.3}) \\
\bottomrule
\end{tabular}}
\caption{Object detection on DOTAv2 and DIOR dataset with mAP, validated on the open-vocabulary object detection task.}
\label{tab:dota_comparison_}
\end{table*}

\section{Training Data Synthesis}
Since the quality of data generated by DMs can vary significantly, we propose a rule-based filtering method, R-Filter, to further refine the generated samples and retain only those that meet predefined quality criteria. Algorithm \ref{alg:alg2} shows training data synthesis with EarthSynth.

\textbf{Condition Sampling.} During the data synthesis stage, conditions $c_i=(m_i,t_i)$ are randomly sampled by category from the training condition set $\mathcal{C} = \{c_i\}_{i=1}^{|\mathcal{C}|}$. We also apply random \textit{rotation}, \textit{scaling}, and \textit{merging}, based on the mask and text prompts, to get more diverse conditions. \textcolor{black}{Figure~\ref{fig:fig4} shows that EarthSynth can generate some unrealistic logical scenes controlled by different text prompts.}

\textbf{Label Mapping.} Different label mapping functions $\mathcal{F}$ are employed for different downstream tasks. For scene classification, the category label $y_c$ is directly obtained from the associated text $t_i$. For semantic segmentation, the semantic mask $m_i$ and the category label $y_c$ are derived by extracting class mappings from the corresponding mask. For object detection, bounding boxes are generated by extracting the contour features of the masks by using the Ramer-Douglas-Peucker~\cite{cao2022automatic} algorithm.

\textbf{Rule-based Filtering.} We propose R-Filter, a rule-based method that uses CLIP scores to evaluate \{image $x$, object $x_{obj}$, background $x_{bg}$\} triplets by computing overall, object-specific, and background scores, as shown in Figure\ref{fig:fig3}. Since both $x_{obj}$ and $x_{bg}$ are related to the label $y$, we retain samples with high overall or object-specific scores for training downtown models by setting the CLIP score threshold $S_0$.

\section{Experiment}\label{sec:sec6}
In this section, we evaluate EarthSynth on scene classification, object detection, and semantic segmentation, including performance analysis, ablation studies, and visualization analysis.

\textbf{EarthSynth-180K.} \textcolor{black}{The diffusion model is trained on cross-satellite and multi-sensor data to enhance generation diversity and improve object modeling across varying observation conditions.} EarthSynth-180K is built from OEM~\cite{xia2023openearthmap}, LoveDA~\cite{wang2021loveda}, DeepGlobe~\cite{demir2018deepglobe}, SAMRS~\cite{wang2023samrs}, and LAE-1M~\cite{pan2025locate} datasets \textcolor{black}{from different satellites and sensors}, and enhanced with mask and text prompts. We leverage many pixel-level mask annotations and semantic-level texts as prompts for DMs. By applying Random Cropping Strategy and Category-Augmentation Strategy to the EarthSynth-180K dataset, we obtain about 180K high-quality triplets consisting of images, semantic masks, and texts. This dataset has a wide range of categories and is labeled with semantic segmentation, which can be used to improve object detail reconstruction and category understanding in DMs. More details can be found in the Appendix.

\textbf{Evaluation.} We use multiple tasks to evaluate data generation capability, including scene classification~\cite{cheng2017remote}, object detection~\cite{zou2023object}, and semantic segmentation~\cite{guo2018review}. We construct an evaluation task that progresses from coarse-grained to fine-grained levels, spanning from image-level to pixel-level, to assess the generalization capability of synthetic data.

\textbf{Experiment Setup.} All experiments are performed using four NVIDIA A100 GPUs, and the complete training of EarthSynth requires approximately 4 $\times$ 45  GPU hours. EarthSynth is initialized with the pretrained Stable Diffusion v1-5~\cite{rombach2022high} weights. Training uses mixed precision to improve computational efficiency and reduce memory consumption. A batch size of 8 per device is used, with gradient accumulation over four steps, resulting in an adequate batch size of 32. In the CF-Comp setting, we set $s_0 = 150$, $\alpha_0 = 1$, $\beta_0 = 0.02$, and $\eta_0 = 0.6$. The local constraint $\gamma$ is set to 10. The training runs for 40,000 steps. A constant learning rate of 1e-5 is adopted without any warm-up phase, and gradient clipping with a maximum norm of 1 is applied to ensure training stability. To improve data quality, we use the CLIP-ViT-B/32 model~\cite{radford2021learning} with the CLIP score threshold $S_0$ set to 0.4.

\subsection{Comparative Results}
\begin{wraptable}{r}{0.5\textwidth}
\centering
\small
\setlength{\tabcolsep}{4pt} % 缩小列间距
\renewcommand{\arraystretch}{0.9} % 增加行高
\begin{tabular}{clccc}
\toprule
& \textbf{Method} & \textbf{Data Usage} & \textbf{mAP}  \\
\midrule
\multirow{7}{*}{\rotatebox{90}{1-shot}} 
& Detic~\cite{zhou2022detecting}  & \textit{Real} & 4.1 \\
& DE-ViT~\cite{zhang2023detect}  & \textit{Real} & 14.7 \\
& CD-ViTO~\cite{fu2024cross}  & \textit{Real} & 17.8 \\
& GroundingDINO~\cite{liu2024grounding}  & \textit{Real} & 11.7 \\
& ETS~\cite{pan2025enhancesearchaugmentationsearchstrategy}  & \textit{Real} & 12.7 \\
&+ ControlNet~\cite{zhang2023adding}  & \textit{Synth} & 9.2 \\
&+ ControlNet~\cite{zhang2023adding}  & \textit{Real} + \textit{Synth} & 13.1 \\
\rowcolor{blue!10}
&+ \textbf{EarthSynth (Ours)}  & \textit{Synth} & 9.5 \\
\rowcolor{blue!10}
&+ \textbf{EarthSynth (Ours)}  & \textit{Real} + \textit{Synth} & 13.9 \\
\midrule
\multirow{7}{*}{\rotatebox{90}{5-shot}} 
& Detic~\cite{zhou2022detecting}  & \textit{Real} & 12.1 \\
& DE-ViT~\cite{zhang2023detect}  & \textit{Real} & 23.4 \\
& CD-ViTO~\cite{fu2024cross}  & \textit{Real} & 26.9 \\
& GroundingDINO~\cite{liu2024grounding}  & \textit{Real} & 27.7 \\
& ETS~\cite{pan2025enhancesearchaugmentationsearchstrategy}  & \textit{Real}  & 29.3 \\
& + ControlNet~\cite{zhang2023adding}  & \textit{Synth}  & 23.0 \\
& + ControlNet~\cite{zhang2023adding}  & \textit{Real} + \textit{Synth} & 33.9 \\
% \rowcolor{gray!20} & (EarthSynth) ETS~\cite{pan2025enhancesearchaugmentationsearchstrategy}  & \textit{Synth} & 23.6 \\
\rowcolor{blue!10} & + \textbf{EarthSynth (Ours)}  & \textit{Synth} & 23.6 \\
\rowcolor{blue!10}
& + \textbf{EarthSynth (Ours)} & \textit{Real} + \textit{Synth} & 34.1 \\
\midrule
\multirow{7}{*}{\rotatebox{90}{10-shot}} 
& Detic~\cite{zhou2022detecting}  & \textit{Real}  & 15.4 \\
& DE-ViT~\cite{zhang2023detect}  & \textit{Real} & 25.6 \\
& CD-ViTO~\cite{fu2024cross}  & \textit{Real} & 30.8 \\
& GroundingDINO~\cite{liu2024grounding}  & \textit{Real} & 36.4 \\
& ETS~\cite{pan2025enhancesearchaugmentationsearchstrategy}  & \textit{Real} & 37.5 \\
& + ControlNet~\cite{zhang2023adding}  & \textit{Synth} & 25.3 \\
& + ControlNet~\cite{zhang2023adding}  & \textit{Real} + \textit{Synth} & 40.2 \\
\rowcolor{blue!10}
& + \textbf{EarthSynth (Ours)}  & \textit{Synth} & 26.4 \\
\rowcolor{blue!10}
&+ \textbf{EarthSynth (Ours)}  & \textit{Real} + \textit{Synth} & 40.7 \\
\bottomrule
\end{tabular}
\caption{The few-shot detection results on the DIOR dataset.}
\vspace{-0.1in}
\label{tab:d1d2d3}
\end{wraptable}
We evaluate downstream remote sensing tasks in open-vocabulary scenarios, including scene classification, object detection, and semantic segmentation. Training data is synthesized using \textit{remote sensing-specific methods} trained on RSICD~\cite{lu2017exploring} Txt2Img-MHN~\cite{xu2023txt2img}, CRS-Diff~\cite{tang2024crs}, and \textit{baseline methods} trained on EarthSynth-180K, Stable Diffusion~\cite{rombach2022high}, InstanceDiffusion~\cite{wang2024instancediffusion}, ControlNet~\cite{zhang2023adding}. We adopt open-vocabulary methods, CLIP~\cite{radford2021learning}, GroundingDINO~\cite{liu2024grounding}, and GSNet~\cite{ye2025towards} for downtown evaluation method. \textit{Real} denotes real-world data, and \textit{Synth} refers to diffusion-generated data. \textcolor{black}{Note that most existing remote sensing diffusion models are based on the above architectures, lack category control, and are not optimized for open-vocabulary understanding tasks.  More detailed analyses are provided in the Appendix.}

\textbf{Scene Classification.} We uniformly generated 10 images per category and averaged the results over three runs for evaluation. Table\ref{tab:rsicd_comparison_} presents the comparative results of \textcolor{black}{CLIP-based} scene classification regarding Top-1 and Top-5 accuracy on RSICD~\cite{lu2017exploring} and DIOR~\cite{li2020object} datasets. RSICD* is a subset of the original RSICD dataset containing 11 classes, used to align different dataset settings of different methods that do not include the RSICD dataset. Compared to VAE-based and GAN-based methods, diffusion-based approaches exhibit a clear advantage in generation quality. On RSICD, our method ranks second in Top-1 accuracy and first in Top-5 accuracy for classification. It also outperforms the baseline ControlNet by 6.3. Our method achieves superior scene classification performance on DIOR, indicating its ability to generate images with higher classification scores. These results suggest the potential of our approach in downstream tasks.

\textbf{Object Detection.}
We evaluate the effectiveness of our proposed method on object detection by training the GroundingDINO model \textcolor{black}{on DOTAv2 or DIOR dataset} with 256 synthetic images per category. Table~\ref{tab:dota_comparison_} experiments on the DOTAv2~\cite{xia2018dota} and DIOR~\cite{li2020object} datasets show that our approach outperforms ControlNet, achieving improvements of 2.1 and 0.3 compared to training solely on real data. We also experimented with a few-shot object detection. Table~\ref{tab:d1d2d3} shows that ETS~\cite{pan2025enhancesearchaugmentationsearchstrategy} consistently outperforms close-source baselines across all shot settings. ETS achieves 37.5 mAP in the 10-shot setting, surpassing Detic at 15.4 and DE-ViT at 25.6. Introducing synthetic data leads to notable improvements, especially when combined with real data. In the 5-shot setting, ETS improves from 29.3 using only real data to 33.9 with ControlNet-generated data and 34.1 with EarthSynth-generated data. \textcolor{black}{These results show that adding synthetic data from EarthSynth can improve open-vocabulary object detection.}

\textbf{Semantic Segmentation.} We generated 256 synthetic images for each category and \textcolor{black}{combined them with the real EarthSynth-180K dataset} to train the open-vocabulary segmentation method GSNet. Table~\ref{tab:miou_comparison} reports the mIoU results on four unseen semantic segmentation datasets: Potsdam~\cite{ISPRS2013}, FloodNet~\cite{rahnemoonfar2021floodnet}, FAST~\cite{wang2023samrs}, and FLAIR~\cite{garioud2023flair}. The baseline GSNet model achieves solid performance using only real data, particularly on the Potsdam dataset, and incorporating synthetic data from ControlNet results in mixed outcomes, with performance improvements on FloodNet by 5.4, FAST by 0.9, and FLAIR by 1.1. Still, a degradation of Potsdam by 5.2, suggesting potential issues related to multi-source data alignment. In contrast, EarthSynth consistently improves performance across all datasets, with gains of 2.1 on Potsdam, 10.5 on FloodNet, 0.6 on FAST, and 2.3 on FLAIR. These results demonstrate the effectiveness of EarthSynth-generated data in enhancing semantic segmentation under diverse scenarios.

\begin{table*}[t]
\centering

\resizebox{\linewidth}{!}{
\begin{tabular}{l l c c c c}
\toprule
\multirow{2}{*}{\textbf{Method}} & \multirow{2}{*}{\textbf{Data Usage}} & \multicolumn{4}{c}{\textbf{Semantic Segmentation}} \\
\cmidrule(lr){3-6}
&  & Potsdam  & FloodNet  & FAST  & FLAIR  \\
\midrule
Base GSNet & \textit{Real} & 40.6 & 33.9 & 16.8 & 19.3 \\
+ StableDiffusion~\cite{rombach2022high} & \textit{Real} + \textit{Synth} & - & - & - & - \\
+ ControlNet~\cite{zhang2023adding} & \textit{Real} + \textit{Synth} & 35.4 (\textcolor{green}{-5.2}) & 39.3 (\textcolor{blue}{+5.4}) & \textbf{17.7} (\textcolor{blue}{+0.9}) & 20.4 (\textcolor{blue}{+1.1}) \\
% \rowcolor{gray!20}EarthSynth + GSNet & \textit{Real} + \textit{Synth} & \textbf{42.7} (\textcolor{blue}{+2.1}) & \textbf{44.4} (\textcolor{blue}{+10.5}) & 17.4 (\textcolor{blue}{+0.6}) & \textbf{21.6} (\textcolor{blue}{+2.3})\\
\rowcolor{blue!10} + \textbf{EarthSynth (Ours)}& \textit{Real} + \textit{Synth} & \textbf{42.7} (\textcolor{blue}{+2.1}) & \textbf{44.4} (\textcolor{blue}{+10.5}) & 17.4 (\textcolor{blue}{+0.6}) & \textbf{21.6} (\textcolor{blue}{+2.3})\\
\bottomrule
\end{tabular}}
\caption{Semantic segmentation on Potsdam, FloodNet, FAST, and FLAIR datasets with mIoU.}
\label{tab:miou_comparison}
\end{table*}

\begin{figure*}[t]
  \centering

  % 图片部分
  \begin{minipage}[c]{0.48\linewidth}
    \centering
    \includegraphics[width=0.9\linewidth]{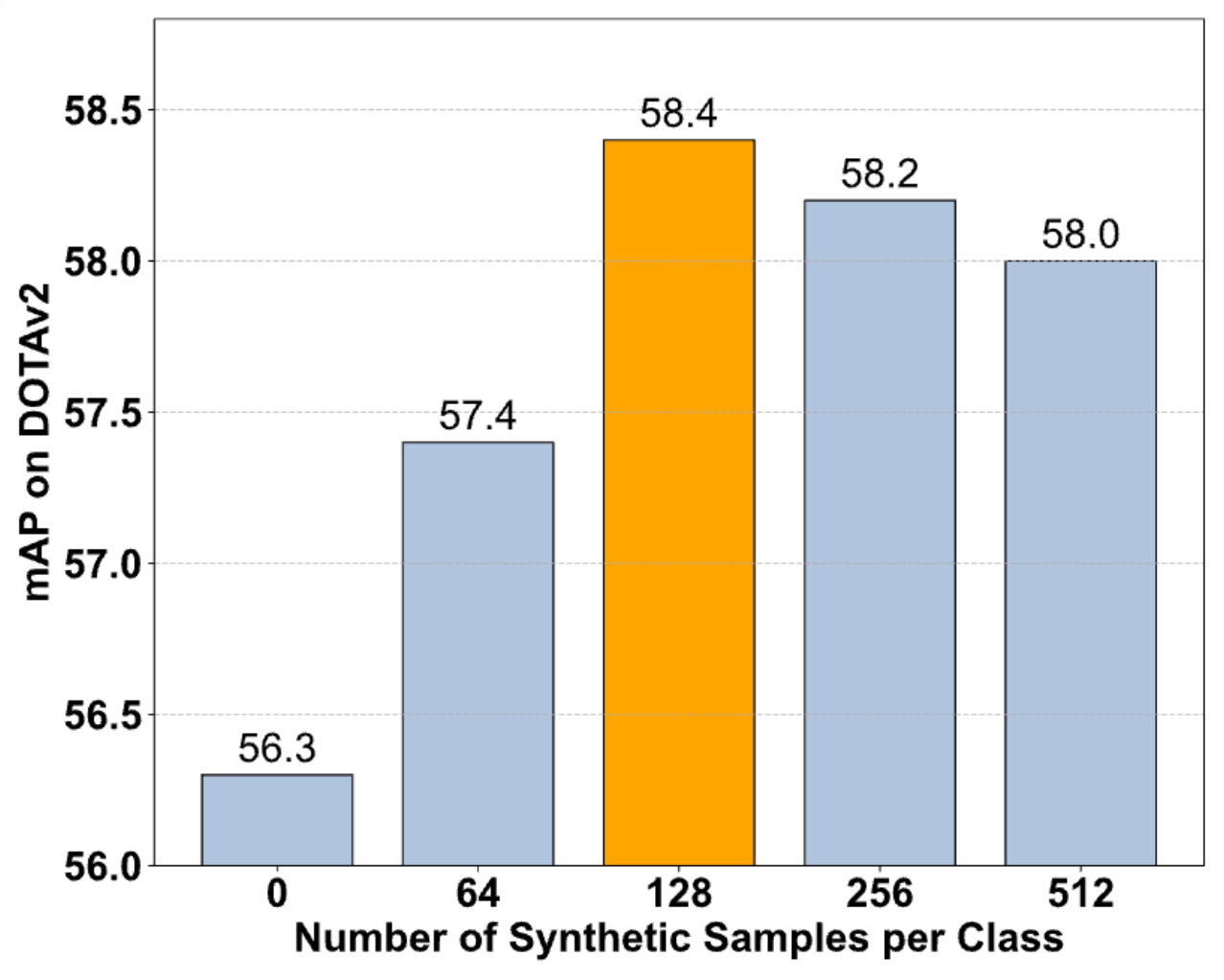} % 可调整为你的图片路径
    \caption{Effect of samples per class on DOTAv2 dataset.}
    \label{fig:fig5}
  \end{minipage}
  \hfill
  % 表格部分
  \begin{minipage}[c]{0.45\linewidth}
    \centering
    \renewcommand{\arraystretch}{1.2}
    \resizebox{0.9\linewidth}{!}{
    \begin{tabular}{lc}
      \toprule
      \textbf{Method}  & \textbf{mAP} \\
      \midrule
      *only real                         & 56.3 \\
      baseline                           & - \\
      + $\mathcal{L}_{\text{local}}$     & 56.5 \\
      + $\mathcal{L}_{\text{local}}$ + R-Filter & 57.4 \\
      + CF-Comp                    & - \\
      + CF-Comp + $\mathcal{L}_{\text{local}}$ & \underline{57.9} \\
      % \rowcolor{gray!20}+ CF-Comp + $\mathcal{L}_{\text{local}}$ + R-Filter & \textbf{58.4} \\
     \rowcolor{blue!10}+ CF-Comp + $\mathcal{L}_{\text{local}}$ + R-Filter & \textbf{58.4} \\
      \bottomrule
    \end{tabular}}
    % \vspace{-0.1in}
    \captionof{table}{Detection performance on DOTAv2 using different modules configurations. “–” is a failure to learn mask-based semantic control for labeled sample generation.}
    \label{tab:dota_results}
    \vspace{-0.2in}
  \end{minipage}
\end{figure*}

\subsection{Ablation Studies}
In this section, we perform the ablation studies of the submodule, sample size, and CF-Comp.

\textbf{Submodule.} We fix the random seed during inference to ensure consistency across comparative experiments, guaranteeing that identical templates are used throughout the evaluation. As shown in Table\ref{tab:dota_results}, object detection performance on DOTAv2 using different submodule configurations. We found that introducing the local loss $\mathcal{L}_{\texttt{local}}$ can learn layout control of the semantic mask. Adding R-Filter further boosts performance to 57.4. Incorporating CF-Comp alongside $\mathcal{L}_{\texttt{local}}$ brings a larger gain, reaching 57.9. The full configuration with CF-Comp, $\mathcal{L}_{\text{local}}$, and R-Filter achieves the best result of 58.4 mAP. CF-Comp and R-Filter effectively retain higher-quality informative samples for downstream tasks. 

\textbf{Sample Size.} Figure~\ref{fig:fig5} shows that performance improves as the number of synthetic samples per class increases, reaching the highest mAP of 58.4 at 128 samples per class. However, further increases lead to marginal declines, suggesting that moderate-scale synthesis is most effective, while excessive data may introduce redundancy or noise.

\textbf{CF-Comp and CLIP of R-Filter.} 
Further ablation studies on CF-Comp and CLIP of R-Filter are in the Appendix.

\subsection{Visualization Analysis}
\begin{figure}[t]
\vskip 0.2in
\begin{center}
\centerline{\includegraphics[width=\columnwidth]{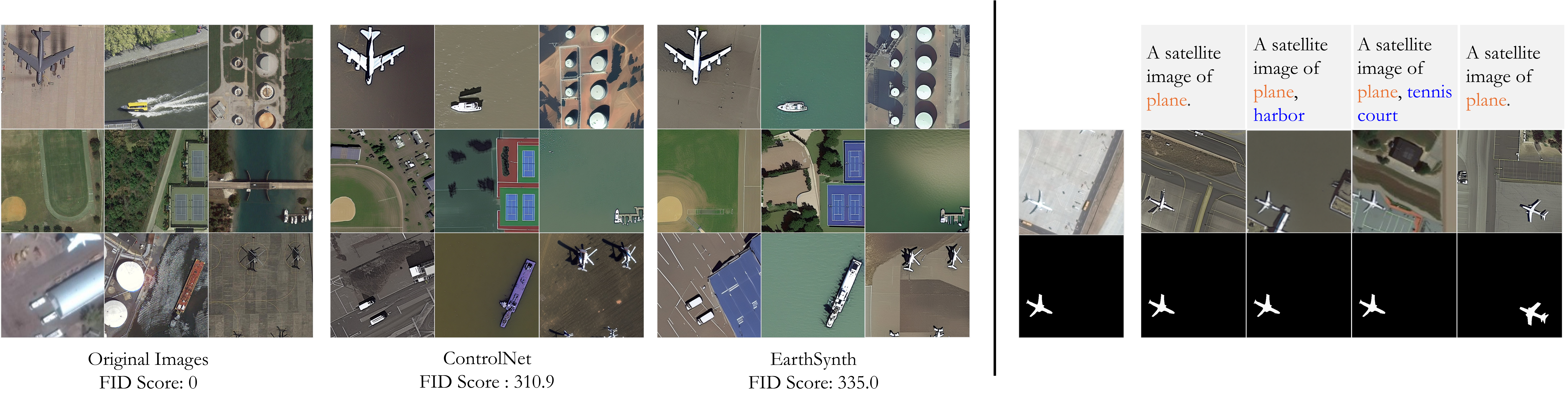}}
\caption{Left: Visualization of synthesis satellite images on DOTAv2 dataset. Right: EarthSynth can generate some unrealistic logical scenes controlled by different text prompts.}
\label{fig:fig4}
\end{center}
\vskip -0.2in
\end{figure}

\textbf{Synthesis.} Figure~\ref{fig:fig4} shows the synthesis satellite images with FID scores~\cite{heusel2017gans} on DOTAv2 dataset. Compared to the baseline ControlNet, EarthSynth in larger distributional shifts and generates higher-quality images. In addition, by combining prompts with transformed masks, we can produce more robust images for downstream tasks.

\begin{figure}[t]
\vskip 0.2in
\begin{center}
\centerline{\includegraphics[width=0.8\columnwidth]{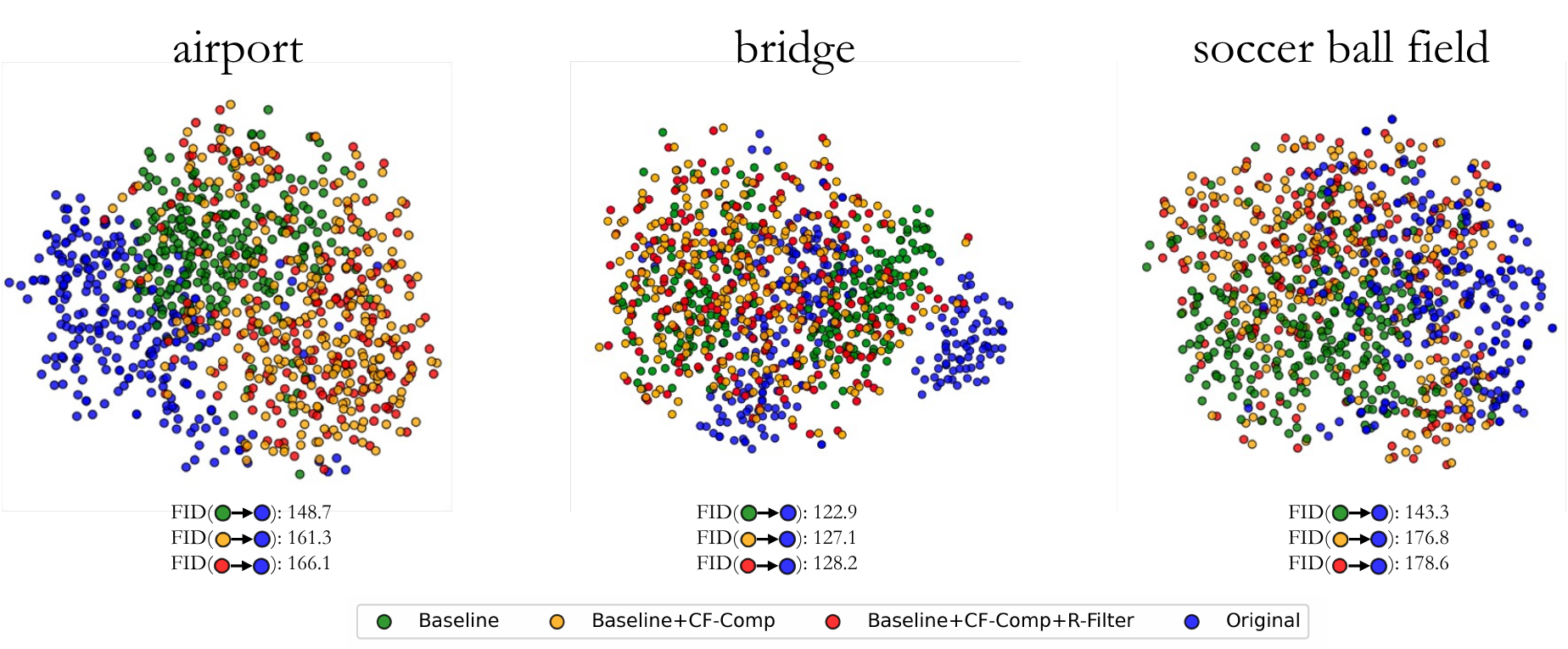}}
\caption{Embedding visualizations for some categories on the EarthSynth-180K dataset.}
\label{fig:fig6}
\end{center}
\vskip -0.2in
\end{figure}

\textbf{Embedding.} Figure~\ref{fig:fig6} shows the t-SNE~\cite{van2008visualizing} distributions and FID scores for three representative categories, illustrating the alignment between synthetic and real data. The use of CF-Comp increases sample diversity but also leads to larger distributional shifts, as indicated by higher FID scores. The introduction of R-Filter also increases this gap, with FID scores consistently increasing across three categories. These results suggest that CF-Comp and R-Filter enhance the semantic diversity and distributional spread of synthetic samples, potentially improving generalization for downstream tasks.

\section{Conclusion}
We propose EarthSynth, a diffusion-based generative foundation model trained on the EarthSynth-180K dataset, which is the first large-scale remote sensing dataset for diffusion training. This work addresses the challenge of poor generalization in task-oriented synthesis for RSI interpretation tasks. To balance layout controllability and category diversity during training, EarthSynth adopts the batch-based CF-Comp strategy, enabling precise layout control for RSI generation. Additionally, it incorporates the R-Filter post-processing method to extract more informative synthesized samples for downstream tasks. EarthSynth is evaluated in open-world scenarios, offering a practical and scalable solution to advance RSI interpretation through synthetic data.

\noindent\textbf{Limitations.} 
Our work focuses on RGB-based remote sensing imagery, excluding standard multispectral data. It also incurs higher training costs, a known challenge in image generation. Further discussion is provided in the Appendix.

% \begin{ack}
% This work was supported by the National Natural Science Foundation of China (42125503, 42430602).
% \end{ack}

% \bibliographystyle{unsrt}
\bibliographystyle{ieeetr}
\bibliography{ref}

\clearpage
\appendix

\section{Technical Appendices and Supplementary Material}
\subsection{More Preliminaries}
\subsubsection{Copy-Paste Augmentation.} Copy-Paste~\cite{ghiasi2021simple} is a data augmentation technique that involves copying objects or regions from one image and pasting them into another to create new composite scenes. $\texttt{Copy-Paste}(x^a_{obj},x^b)$ represents the operation of copying the objects of image $x^a$ to image $x^b$. \textcolor{black}{However, the Copy-Paste introduces compositional artifacts or non-smooth transitions, etc., that alter the statistical properties of the image distribution. These artifacts and transitions can typically be mitigated through the training process of DMs.}
Given two satellite images, $x^a$ and $x^b$ with their masks $m^a$ and $m^b$ and text embddings $t^a$ and $t^b$, the $\texttt{Copy-Paste}(x^a,x^b)$ operation can be defined as follows:
\[
% x' = \texttt{Copy-Paste}(x^a_{\texttt{obj}},x^b_{\texttt{bg}}), \\
% x'' = \texttt{Copy-Paste}(x^b_{\texttt{obj}},x^a_{\texttt{bg}}), \\
x' = x^a + \mathbf{1}_{\{m^a = 0\}} \cdot x^b ,
\]
\[
m' = m^a + \mathbf{1}_{\{m^a = 0\}} \cdot m^b,
\]
\[
t' = t^b + t^a \setminus (t^a \cap t^b),
\]
where $\mathbf{1}_{\{m^a = 0\}}$ denotes an indicator function that returns 1 when $m^a = 0$, and $0$ otherwise. $\texttt{Copy-Paste}(x^a_{obj},x^b)$ represents the operation of copying the objects of image $x^a$ to image $x^b$.

\subsection{EarthSynth-180K Dataset}\label{app:sub1}
\begin{figure}[h]
\vskip 0.2in
\begin{center}
\centerline{\includegraphics[width=0.9\linewidth]{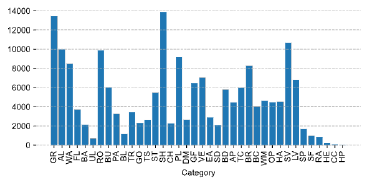}}
\caption{Category distribution of the EarthSynth-180K.}
\label{ex-fig0}
\end{center}
\vskip -0.2in
\end{figure}

EarthSynth-180K is derived from OEM, LoveDA, DeepGlobe, SAMRS, and LAE-1M datasets from different satellites. The satellite sources of the EarthSynth-180K dataset are shown in Table~\ref{tab:satellite_sources}. It is further enhanced with mask and text prompt conditions, making it suitable for training foundation DMs. The EarthSynth-180K dataset is constructed using the Random Cropping and Category-Augmentation strategies. The category distribution of the EarthSynth-180K dataset is presented in Figure~\ref{ex-fig0}, along with the corresponding category-to-abbreviation mapping shown in Table~\ref{tab:category_abbr_map}. Although remote sensing focuses on a limited number of classes, this study validates the feasibility of the method on major classes by expanding the vocabulary, and the limited number of classes does not affect the conclusions.

\subsubsection{Random Cropping Strategy} To standardize the input resolution for the DM, we employ a random cropping strategy to generate 512$\times$512 image patches. For lengths smaller than 1024, resample to 512; the insufficient edge parts are filled with zero values. The same cropping operation is applied to the corresponding semantic masks to maintain spatial consistency. Specifically, if the image size is larger than the target crop size, a random top-left coordinate $(x, y)$ is selected to extract a patch of the desired dimensions. If the image is smaller than the crop size, no cropping is performed, and the region starting from the top-left corner is used directly. To construct textual descriptions for the categories, we employ a template-based approach by appending category names to the phrase ``\textit{A satellite image of [class1], [class2], ...}'', resulting in descriptions such as ``\textit{A satellite image of \underline{swimming pool}, \underline{ship}, \underline{small vehicle, harbor}}''. We construct around 180K triplets of images, textual descriptions, and semantic masks. Note that center-based cropping is not required, as the categories in the dataset are heterogeneous and spatially scattered.

\begin{figure}[t]
% \vskip 0.2in
\begin{center}
\centerline{\includegraphics[width=0.7\linewidth]{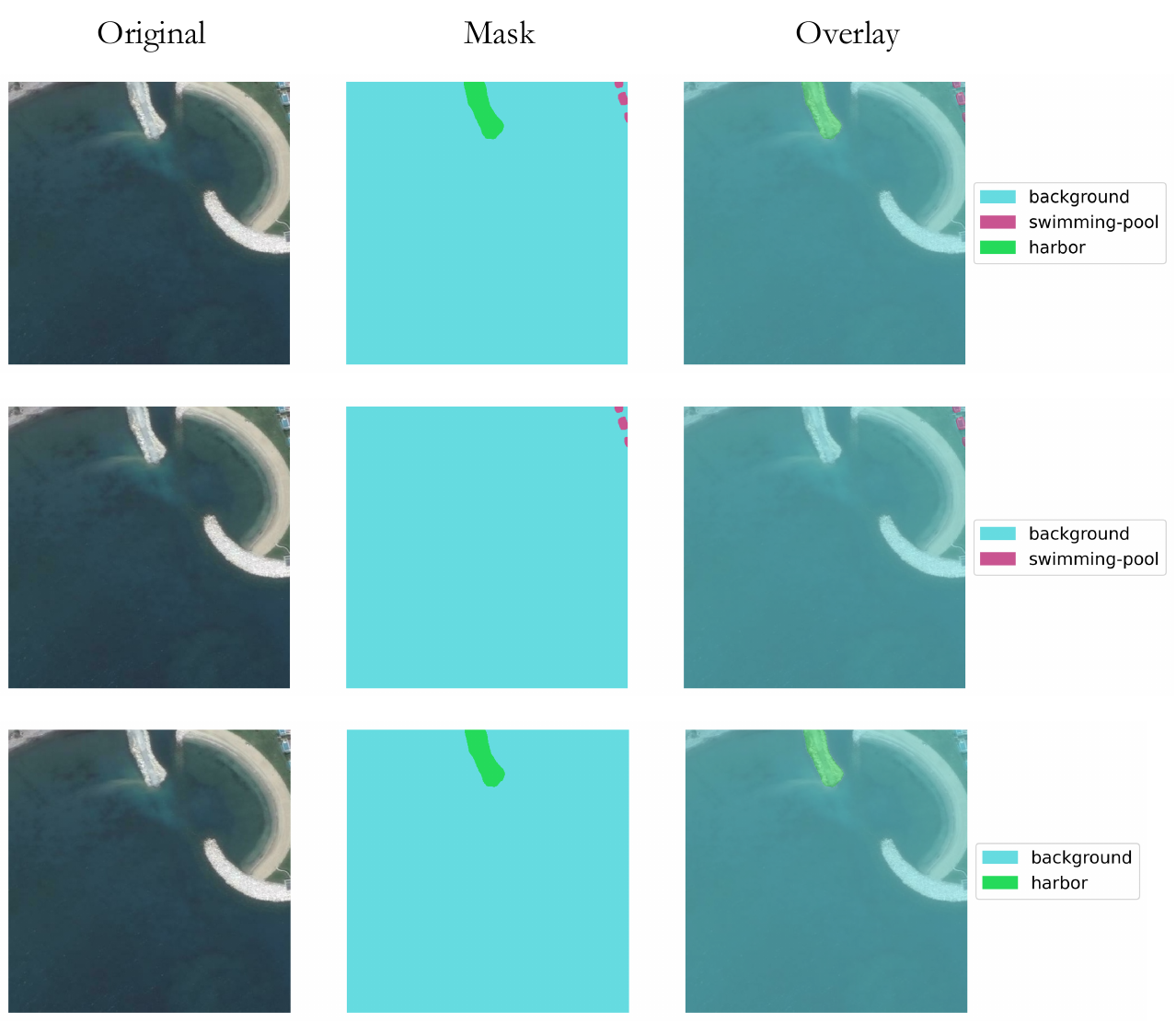}}
\caption{A category-augmentation strategy to construct multiple object-background pairs.}
\label{ex-fig1}
\end{center}
% \vskip -0.2in
\end{figure}

\subsubsection{Category-Augmentation Strategy} We apply data augmentation to each image to enhance the model’s understanding of individual categories and enable fine-grained, category-specific control during generation, creating multiple distinguishable foreground-background pairs. Additionally, this approach helps improve the probability of sample combinations in the batch-based CF-Comp strategy. Figure~\ref{ex-fig1} shows that these augmented triplets are generated from the previously obtained \texttt{image-text-mask} triplets by isolating non-background categories in the original masks and corresponding textual descriptions, resulting in a set of new, category-focused triplets. We obtained approximately 500K paired textual descriptions and semantic masks of varying granularities through category augmentation.

\begin{table*}[t]
\centering
\renewcommand{\arraystretch}{1.2}
\resizebox{0.8\linewidth}{!}{
\begin{tabular}{lll lll lll}
\toprule
Category & Abbr. & 
Category & Abbr. & 
Category & Abbr. \\
\midrule
background & BG & ground track field & GF & vehicle & VE \\
bare land & BL & small vehicle & SV & windmill & WM \\
grass & GR & baseball diamond & BD & expressway service area & EA \\
pavement & PA & tennis court & TC & expresswalltoll station & ET \\
road & RO & roundabout & RA & dam & DM \\
tree & TR & storage tank & ST & golf field & GO \\
water & WA & harbor & HA & overpass & OP \\
agriculture land & AL & container crane & CC & stadium & SD \\
buildings & BU & airport & AP & train station & TS \\
forest land & FL & helipad & HP & large vehicle & LV \\
barren land & BA & chimney & CH & swimming pool & SP \\
urban land & UL & helicopter & HE & bridge & BR \\
plane & PL & ship & SH & soccer ball field & SF \\
basketball court & BC & & & & \\
\bottomrule
\end{tabular}}
\caption{The main category to abbreviation mapping.}
\label{tab:category_abbr_map}
\end{table*}

\begin{table*}[ht]
\centering
\resizebox{\linewidth}{!}{
\begin{tabular}{lcc}
\toprule
Dataset & Data Sources & Sensor Type \\
\midrule
OEM~\cite{xia2023openearthmap} & Existing Benchmark
Dataset, Various Satellite Operators and Agencies & Satellite, Aircraft, and UAV \\
LoveDA~\cite{wang2021loveda} & Google Earth Platform & Satellite \\
DeepGlobe~\cite{demir2018deepglobe} & WorldView-2 & Satellite\\
SAMRS~\cite{wang2023samrs} & Sentinel-1, Sentinel-2, PlanetScope, and others & Satellite, Aircraft, and UAV \\
LAE-1M~\cite{pan2025locate} & Existing Object Detection Datasets, Google Earth Platform &  Satellite, Aircraft, and UAV\\
\bottomrule
\end{tabular}}
\caption{Data sources and sensor type of EarthSynth-180K.}
\label{tab:satellite_sources}
\end{table*}

\subsubsection{Data Sources of EarthSynth-180K} Table~\ref{tab:satellite_sources} summarizes the satellite platforms and corresponding sensor types that contribute to the EarthSynth-180K dataset.

\textbf{OEM.}
The Open Earth Map (OEM) dataset is a global initiative to advance open machine learning-based mapping techniques using remote sensing data. It focuses on extracting semantic and height information, such as land cover maps and digital elevation models (DEMs), to support environmental monitoring, urban planning, and disaster management applications. The dataset comprises 5,000 aerial and satellite images with manually annotated 8-class land cover labels at a 0.25–0.5m ground sampling distance, covering 97 regions from 44 countries across six continents.

\textbf{LoveDA.}
LoveDA dataset is designed for land-cover domain adaptation semantic segmentation. It contains 5,987 high spatial resolution (0.3m) remote sensing images from three cities in China, including urban and rural scenes. The images are sourced from the Google Earth Platform, providing real-world urban and rural remote sensing images for semantic segmentation and unsupervised domain adaptation tasks.

\textbf{DeepGlobe.}
DeepGlobe dataset is part of the DeepGlobe 2018 Satellite Image Understanding Challenge, which includes three public competitions for segmentation, detection, and classification tasks on satellite images. The dataset consists of high-resolution satellite images with a 50cm pixel resolution collected by DigitalGlobe's WorldView series satellites. It is used for road extraction and building detection tasks.

\begin{wrapfigure}{r}{0.5\textwidth}
% \vskip 0.2in
% \begin{center}
\centerline{\includegraphics[width=\linewidth]{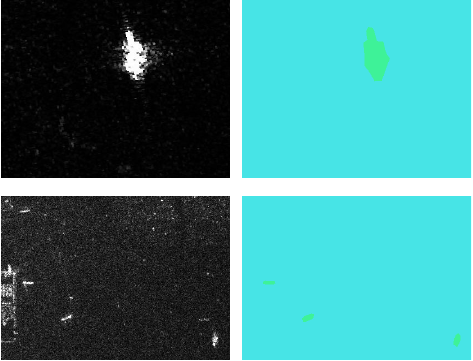}}
\caption{Some images of the EarthSynth-180K dataset are derived from SAR imagery.}
\label{ex-sar}
% \end{center}
\end{wrapfigure}

\textbf{SAMRS.}
SAMRS dataset is a large-scale remote sensing segmentation dataset developed using the Segment Anything Model~\cite{kirillov2023segment}. It leverages existing remote sensing object detection datasets to generate a comprehensive dataset for semantic segmentation, instance segmentation, and object detection tasks. The dataset comprises 105,090 images with 1,668,241 instances, surpassing existing high-resolution remote sensing segmentation datasets in size by several orders of magnitude. It integrates data from various sources, including Sentinel-1, Sentinel-2, and PlanetScope satellites.

\textbf{LAE-1M.}
LAE-1M dataset is a remote sensing object detection dataset with broad category coverage. It is constructed by unifying up to 10 remote sensing datasets to create a comprehensive collection for open-vocabulary object detection tasks. The dataset includes high-resolution optical satellite imagery, though specific satellite sources are not disclosed. We used part of the LAE-1M dataset as expanded semantic diversity. While this work mainly concentrates on optical image understanding and generation, a portion of the ship category in the dataset is derived from SAR imagery, as shown in Figure~\ref{ex-sar}. Due to the scarcity of such data, understanding cross-sensor categories remains a vital aspect to consider.

\begin{table*}[t]
\centering
\resizebox{\linewidth}{!}{
\begin{tabular}{lccc}
\toprule
\multirow{2}{*}{Method} & \multicolumn{3}{c}{Condition Type} \\
\cmidrule(lr){2-4}
 & Semantic-level (class, text) & Region-level (box) & Pixel-level (mask, sketch) \\
\midrule
Text2Earth~\cite{liu2025text2earth}   & \checkmarks & \xmark & \xmark \\
DiffusionSat~\cite{khanna2023diffusionsat} & \checkmarks & \xmark & \xmark \\
GeoSynth~\cite{sastry2024geosynth}     & \checkmarks & \xmark & \checkmarks \\
\midrule
CRS-Diff~\cite{tang2024crs}     & \checkmarks & \xmark & \checkmarks \\
SatSynth~\cite{toker2024satsynth}     & \checkmarks & \xmark & \xmark \\
AeroGen~\cite{tang2024aerogen}      & \checkmarks & \checkmarks & \xmark \\
MMO-IG~\cite{yang2025mmo}      & \checkmarks & \checkmarks & \xmark \\
\rowcolor{blue!10}
\textbf{EarthSynth (Ours)}   & \checkmarks & \xmark & \checkmarks \\
\bottomrule
\end{tabular}}
\caption{Comparison of remote sensing diffusion methods based on different levels of prompts.}
\label{tab:prompt_comparison}
\end{table*}

\begin{table*}[t]
\centering
\resizebox{\linewidth}{!}{
\begin{tabular}{clccc}
\toprule
Type & Method & Training Dataset & FID Score & CLIP Score \\
\midrule
\multirow{10}{*}{\rotatebox{90}{\textit{Single}}}
& StableDiffusion~\cite{rombach2022high} & RSICD      & 103.4 & 26.1  \\
& StableDiffusion~\cite{rombach2022high} & DIOR     & 228.1 & 26.2   \\
& InstanceDiffusion~\cite{wang2024instancediffusion} & RSICD     & 138.6 & 24.7    \\
& Text2Earth~\cite{liu2025text2earth} & Git-10M   & 24.5* & - \\
& DiffusionSat~\cite{khanna2023diffusionsat} & fMoW & 15.8* & 17.2* \\
& GeoSynth~\cite{sastry2024geosynth} & US Cities      & 12.3* / 171.1 & 30.3* / 25.0 \\
& SatSynth~\cite{toker2024satsynth} & iSAID, LoveDA, OEM     & -    & -    \\
& MMO-IG~\cite{yang2025mmo} & DIOR       & 34.5* & -    \\
& AeroGen~\cite{tang2024aerogen} & HRSC,DIOR    & 38.6* & -    \\
\midrule
\multirow{4}{*}{\rotatebox{90}{\textit{Multi}}}
% & GeoSSD~\cite{zhang2024rs5m} & (RS5M)      & 170.7 & 26.3    \\
& CRS-Diff~\cite{tang2024crs} & RSICD, fMoW     & 50.7* / 107.0 & 20.3* / 23.6  \\
& InstanceDiffusion~\cite{wang2024instancediffusion} & RSICD, RSITMD, UCMerced      & 123.1 & 25.0   \\
& ControlNet~\cite{zhang2023adding} & EarthSynth-180K     & 183.8 & 25.9    \\
& \textbf{EarthSynth (Ours)} & EarthSynth-180K   & 198.7    & 26.1 \\
\bottomrule
\end{tabular}}
\caption{Comparison of FID and CLIP scores trained on single-source (\textit{Single}) and multi-source \textit{Multi} data. We calculated the FID score between the generated images and the RSICD dataset. * is from the original papers.}
\label{tab:fid_clip_scores}
\end{table*}

\subsection{Remote Sensing Diffusion Models}
Table~\ref{tab:prompt_comparison} compares remote sensing diffusion models across three prompt levels: semantic, region, and pixel. Most methods support only semantic-level prompts, offering global but coarse control. Some models introduce region-level prompts using bounding boxes to enhance spatial precision. Pixel-level prompts, such as masks and sketches, provide the most detailed control and are used by CRS-Diff, GeoSynth, and EarthSynth. EarthSynth uniquely combines semantic and pixel-level prompts, enabling high-level semantics and fine-grained spatial guidance. This reflects a shift toward more precise and controllable image generation.

\textbf{High-Resolution Synthesis}
Text2Earth, DiffusionSat, and GeoSynth are representative methods to generate high-resolution satellite imagery. These models leverage diffusion-based generative frameworks or text-to-image architectures to reconstruct fine-grained spatial details, often guided by auxiliary inputs such as text descriptions, semantic maps, or multi-modal signals. Their primary applications lie in image restoration, super-resolution, cloud removal, and spectral enhancement, which are essential for improving satellite data's visual and analytical quality in scientific and operational settings.

\begin{table*}[t]
\centering
\renewcommand{\arraystretch}{1.2}
\resizebox{0.8\linewidth}{!}{
\begin{tabular}{lccc}
\toprule
Method & Visual Quality & Semantic Richness & Overall Score \\
\midrule
ControlNet & 79.51 & 23.76 & 51.64 \\
\rowcolor{blue!10}
\textbf{EarthSynth (Ours))} & 80.19 & 25.95 & 53.07 \\
\bottomrule
\end{tabular}}
\captionof{table}{Image scoring results on 100 diffusion-generated images using GPT-4.}
\label{tab:evaluation-results}
\end{table*}

\textbf{Task-Oriented Synthesis}
CRS-Diff, SatSynth, AeroGen, and MMO-IG are designed with downstream utility, focusing on generating synthetic data tailored for specific tasks such as land cover classification, object detection, and change detection. These models incorporate task-specific priors, including class labels, semantic layouts, or instance-level masks, to guide the generation process. By aligning the synthesized data with the needs of target tasks, these methods enhance model generalization in low-resource scenarios, enable domain adaptation, and facilitate pretraining or fine-tuning of models in remote sensing applications.

\subsection{More Experiments}\label{app:sec3}

\subsubsection{Comparison of FID and CLIP scores}
Table~\ref{tab:fid_clip_scores} compares fake data from different models using FID and CLIP scores. FID measures distribution distance to all images of the RSICD dataset, while CLIP evaluates semantic alignment. The results reveal a mismatch between FID and CLIP metrics across models. For example, GeoSynth shows low FID (12.3) and high CLIP (30.3), indicating strong visual and semantic quality. In contrast, StableDiffusion on RSICD or DIOR has a high FID (103.4, 228.1) but is similar to CLIP on EarthSynth (26.1). InstanceDiffusion improves FID with multi-source data, yet its CLIP score stays nearly the same (25.0 vs. 24.7), underscoring a gap between visual fidelity and semantic alignment. This suggests that FID alone is not reliable for assessing task-specific data quality.

\subsubsection{Multi-modal LLM for Image Quality Evaluation}

\textbf{Image Scoring based on Multi-modal LLM.}
We use GPT-4 as a text-based tool to evaluate the generated images, since it can't analyze pixels directly. We extract basic data like resolution, color mode, and estimated visual complexity based on color count, then turn this into a descriptive prompt that summarizes the image’s main features. Our prompt consists of two components: (1) a system prompt that directs GPT-4 as an expert evaluator capable of judging image quality and semantic richness based only on textual input. The system prompt states: \textit{``You are an expert image evaluator who assesses image quality and semantic richness from detailed descriptions alone. You do not require access to the actual image; provide a numeric score from 0 to 100 based solely on the description.''} (2) a user prompt that supplies the image description and explicitly instructs GPT-4 to respond in the exact format: \textit{``Score: X. Reason: \textless brief explanation\textgreater''}, where \textit{X} is an integer between 0 and 100. The prompt explicitly forbids disclaimers such as ``I am an AI and cannot view images.''

\begin{lstlisting}[
  backgroundcolor=\color{lightgray},  % 设置背景色
  frame=single,                      % 添加边框
  rulecolor=\color{black},            % 边框颜色
  xleftmargin=10pt,                   % 左侧边距
  xrightmargin=10pt,                  % 右侧边距
  aboveskip=10pt,                     % 上方间距
  belowskip=10pt,                     % 下方间距
  breaklines=true,                    % 允许自动换行
  basicstyle=\ttfamily\footnotesize    % 设置字体样式和大小
]
You are an expert image evaluator. Based ONLY on the description below,
rate the image on two aspects:
1. Visual Quality (integer score 0-100)
2. Semantic Richness (integer score 0-100)

Image description: <image description text>

Respond ONLY with exactly the following format and nothing else:
Visual Quality Score: <integer 0-100>
Semantic Richness Score: <integer 0-100>
Reason: <one brief sentence>
\end{lstlisting}

Image scoring is performed along two key dimensions: (1) visual quality, encompassing factors such as image clarity, resolution, and compositional coherence; and (2) semantic richness, including the quantity of meaningful elements, scene complexity, and the depth of emotional or narrative content. This structured prompt formulation promotes consistent and parsable outputs, reducing subjective variance. Numeric scores are extracted via regular expression matching, and average scores are computed over the entire image set to yield final metrics. The prompts are listed below:
\begin{table*}[t]
\centering
\renewcommand{\arraystretch}{1.2}
\resizebox{0.7\linewidth}{!}{
\begin{tabular}{lcc}
  \toprule
  Method & CLIP Train Data & Top1 Acc \\
  \midrule
  Baseline (EarthSynth)      &       -          & 49.07 \\
  + RemoteCLIP   & RET-3 + DET-10 + SEG-4  & 48.13 \\
  \rowcolor{blue!10}
  + CLIP  & Web & 50.47 \\
  \bottomrule
\end{tabular}}
\captionof{table}{CLIP-based scene classification performance on DOTA-v2 using different module configurations.}
\label{tab:dota_results_zsd}
\end{table*}

\textbf{Experimental Setup and Results.} 
We evaluated 100 generated images for each method. The assessment was performed using GPT-4, which rated the images along two dimensions: \textit{Visual Quality}, reflecting the perceptual fidelity and aesthetic appeal, and \textit{Semantic Richness}, indicating the depth and variety of semantic content present in the image. The \textit{Overall Average Score} was computed as the arithmetic mean of the two individual scores.

As shown in Table~\ref{tab:evaluation-results}, EarthSynth achieved a slightly higher visual quality score (80.19) than ControlNet (79.51), indicating marginally better perceptual quality. EarthSynth produced images with greater semantic richness, scoring 25.95 compared to 23.76 from ControlNet. Consequently, ControlNet achieved a higher overall average score of 53.07, compared to 51.64 for EarthSynth. These results suggest EarthSynth performs better at both visual quality and semantic richness. We believe its scoring results can only serve as a reference and are intended to provide one of the methods for diversified image quality evaluation.

\begin{figure}[h]
% \vskip 0.2in
% \begin{center}
\centerline{\includegraphics[width=0.8\linewidth]{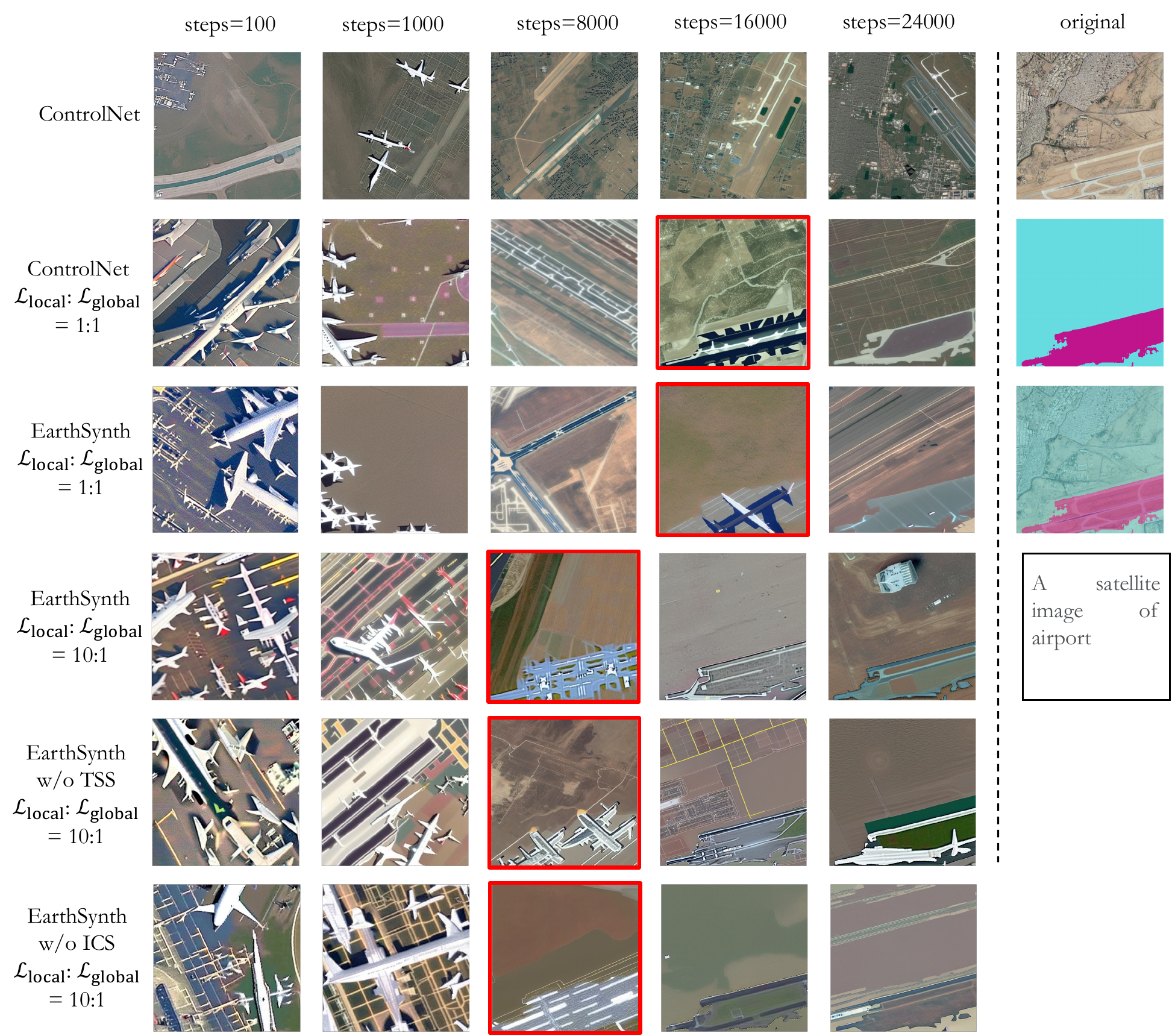}}
\caption{EarthSynth over time-step training process.}
\label{ex-fig3}
% \end{center}
% \vskip -0.2in
\end{figure}

\subsubsection{Ablation Studies}
\begin{figure}[t]
% \vskip 0.2in
% \begin{center}
\centerline{\includegraphics[width=0.8\linewidth]{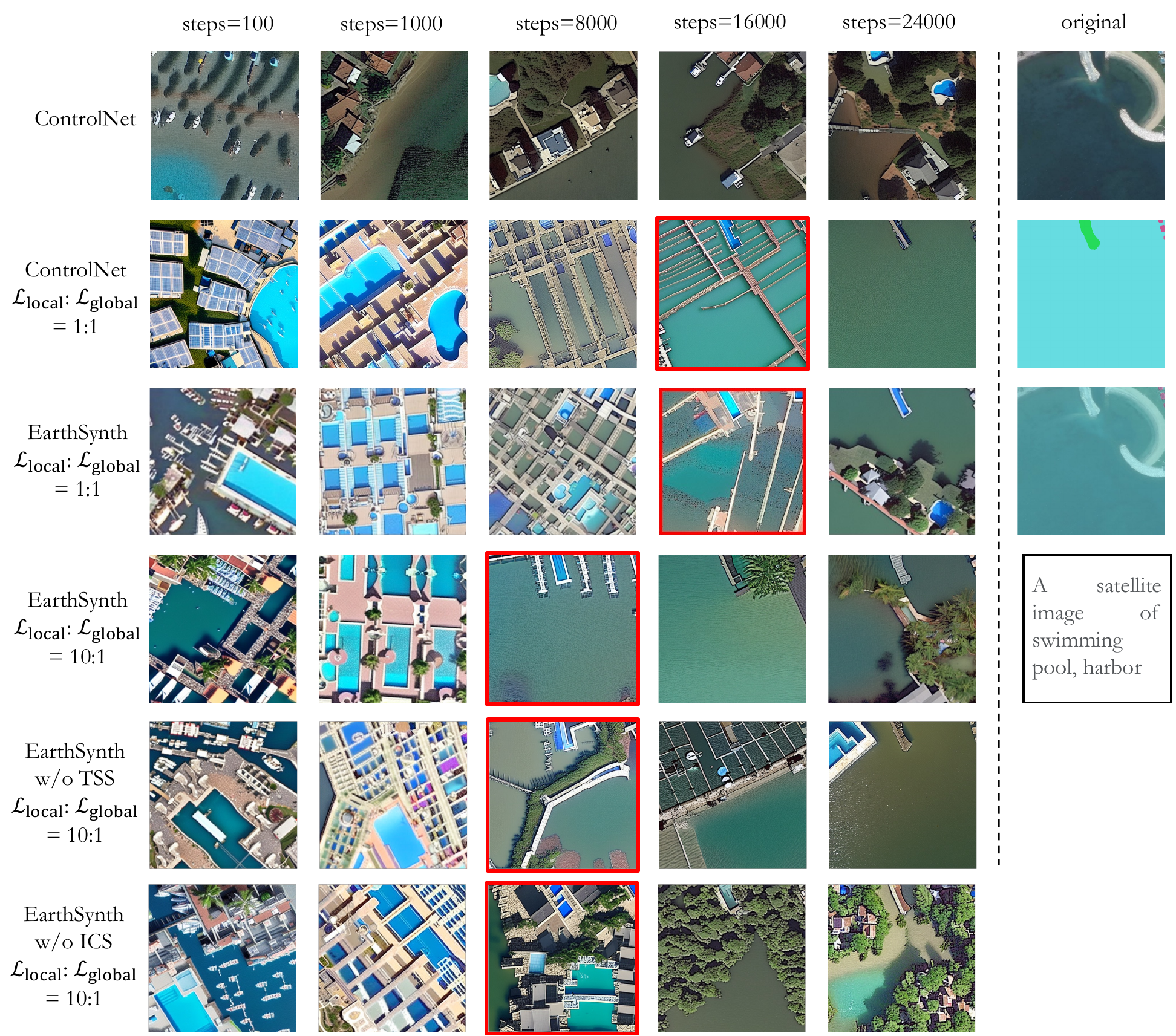}}
\caption{EarthSynth over time-step training process.}
\label{ex-fig2}
% \end{center}
% \vskip -0.2in
\end{figure}

\textbf{Local Loss.} We qualitatively analyze the role of local constraints by visualizing the training process of the EarthSynth model. As shown in Figure~\ref{ex-fig2} and Figure~\ref{ex-fig3}, the loss function is defined as,
\[
\mathcal{L} = \mathcal{L}_{\texttt{global}} + \gamma \mathcal{L}_{\texttt{local}}.
\]
Setting $\gamma = 10$ accelerates convergence and helps the model capture semantic mask information better. Compared to not using $\mathcal{L}_{\texttt{local}}$, incorporating it enables more effective layout control. We found it challenging to learn layout control without local constraints on the mask. For comparison fairness, ControlNet with $\mathcal{L}_{\texttt{local}}$ is used in object detection and semantic segmentation.

\textbf{CF-Comp.} As illustrated in Figure~\ref{ex-fig2} and Figure~\ref{ex-fig3}, we qualitatively analyze the impact of local constraints by visualizing the training process of the EarthSynth model. A visual ablation study on ICS and TSS further highlights their complementary roles within our framework. Specifically, incorporating ICS encourages the model to generate more diverse and novel compositional combinations, enhancing creativity and variety. In contrast, applying TSS fosters the generation of semantically coherent and realistic compositions, improving overall plausibility. These findings also validate the effectiveness of our proposed EarthSynth with CF-Comp strategy, demonstrating its ability to balance novelty with semantic fidelity.

\textbf{CLIP of R-Filter.}  Table~\ref{tab:dota_results_zsd} presents the Top-1 accuracy of different configurations of the EarthSynth model on the DOTA-v2 scene classification task. The baseline EarthSynth model achieves an accuracy of 49.07. When integrated with RemoteCLIP, performance slightly drops to 48.13, suggesting that RemoteCLIP may not effectively enhance EarthSynth in this context, possibly due to misalignment with remote sensing imagery. In contrast, incorporating the standard CLIP module leads to a significant improvement, reaching the highest accuracy of 50.47. This result highlights CLIP’s strong capability to boost scene classification performance. Also, this reacts to the semantic bias that RemoteCLIP carries in the remote sensing domain, which is not well used for filtering.

\begin{figure}[t]
% \vskip 0.2in
% \begin{center}
\centerline{\includegraphics[width=0.8\linewidth]{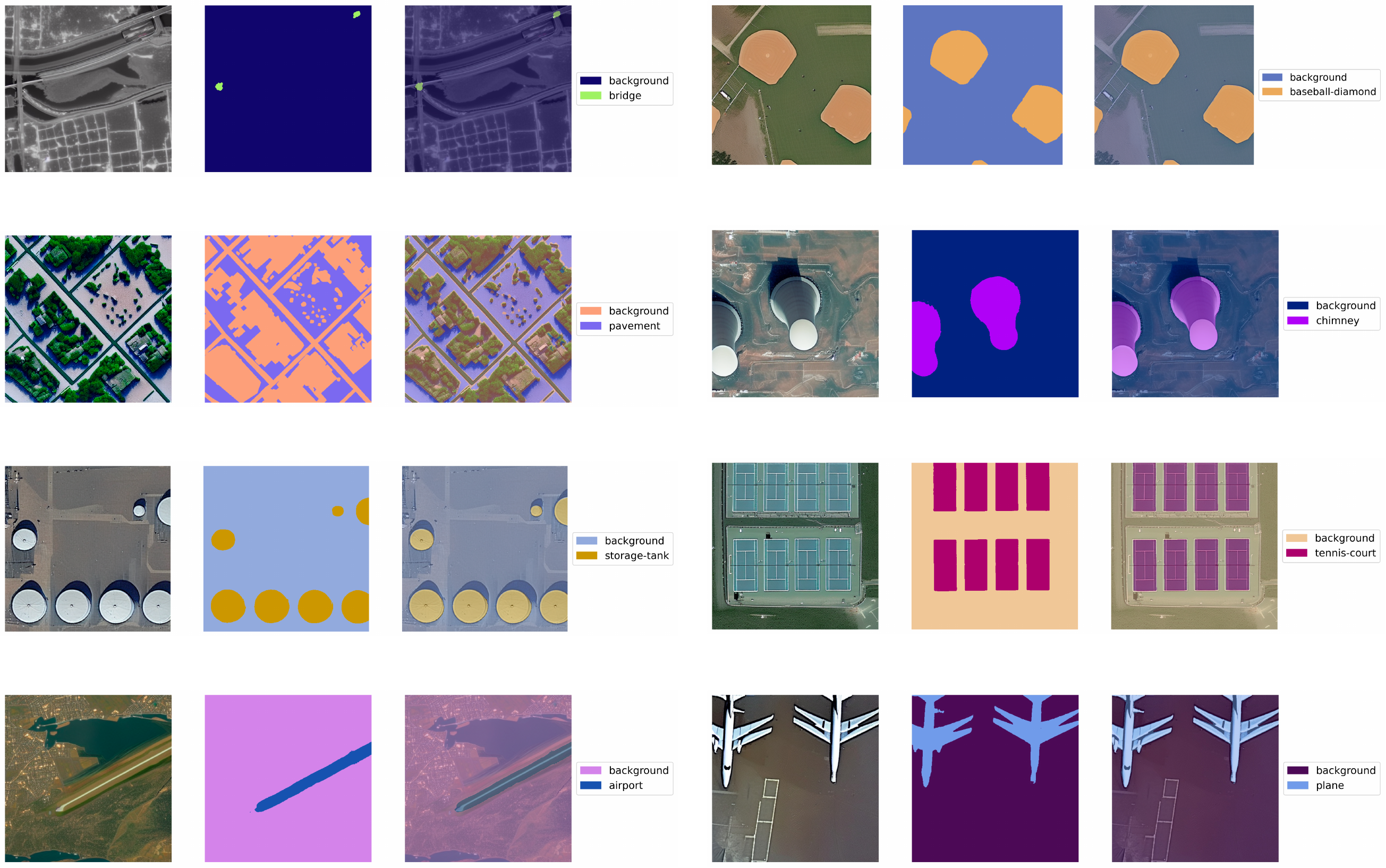}}
\caption{Examples of generated segmentation data.}
\label{ex-fig4}
% \end{center}
% \vskip -0.2in
\end{figure}

\subsection{More Visualization}

\subsubsection{Examples of EarthSynth-generated Data}
\begin{figure}[t]
% \vskip 0.2in
% \begin{center}
\centerline{\includegraphics[width=0.8\linewidth]{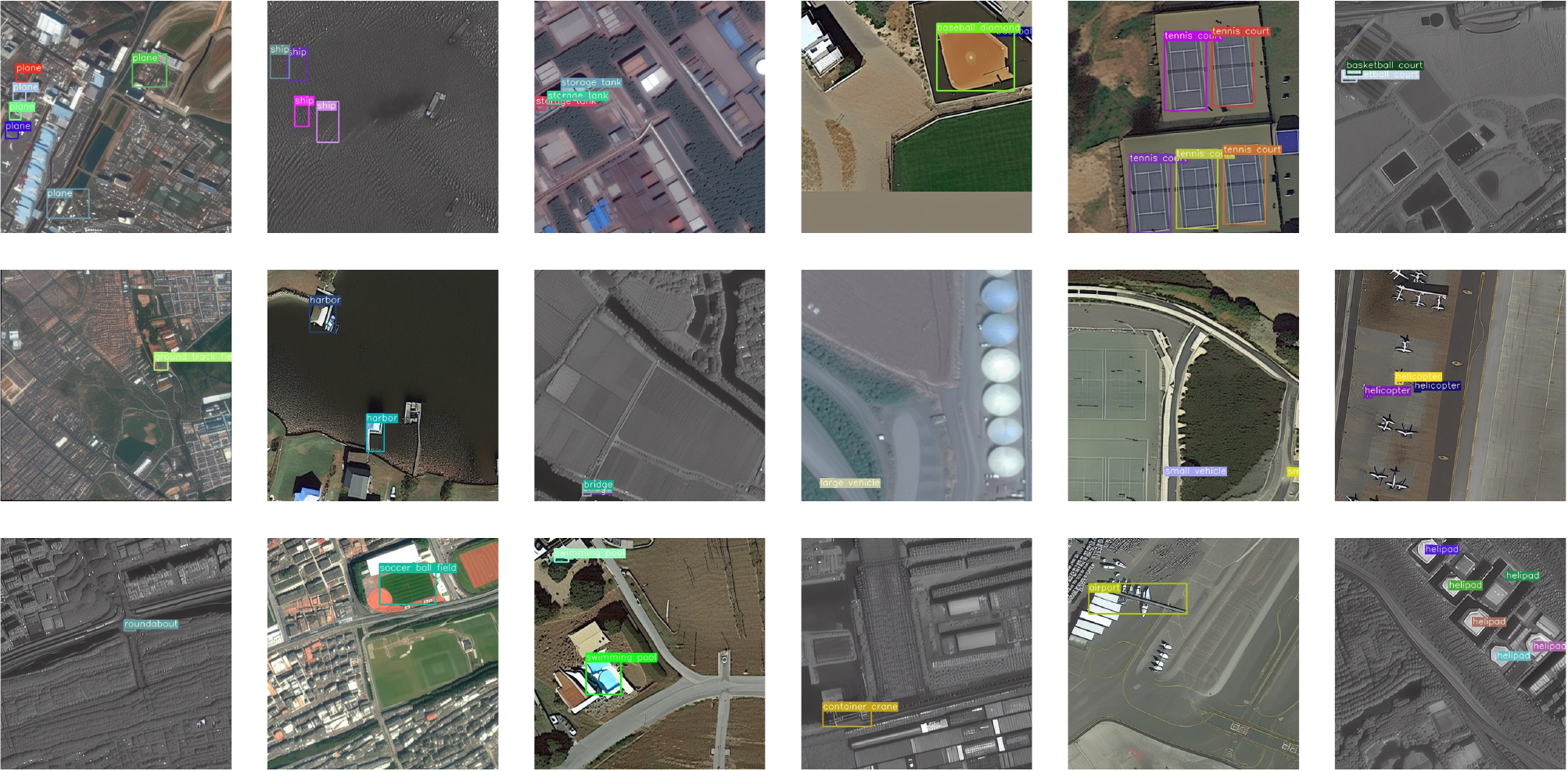}}
\caption{Examples of generated object detection data. This includes the ability to generate data from different satellites.}
\label{ex-fig6}
% \end{center}
% \vskip -0.2in
\end{figure}

Figure~\ref{ex-fig4} and Figure~\ref{ex-fig6} illustrate representative examples of the synthetic semantic segmentation and object detection datasets generated using the proposed EarthSynth framework. These visualizations show that the semantic segmentation labels exhibit higher precision and consistency across diverse land cover types. This can be attributed to the pixel-level supervision involved in the segmentation process, which enables more accurate delineation of object boundaries. Object detection annotations are less reliable because they rely on post-processing techniques like edge detection and bounding box generation. These methods often introduce noise or result in misalignment with the actual object extents, thereby reducing the overall annotation quality. This contrast highlights the relative robustness of the segmentation outputs and suggests that EarthSynth is particularly well-suited for applications where spatial accuracy and detailed contextual information are essential.

\subsubsection{Comparison of Generated Images from Remote Sensing Diffusion Models}
\begin{wrapfigure}{r}{0.55\textwidth}
% \vskip 0.2in
\begin{center}
\centerline{\includegraphics[width=\linewidth]{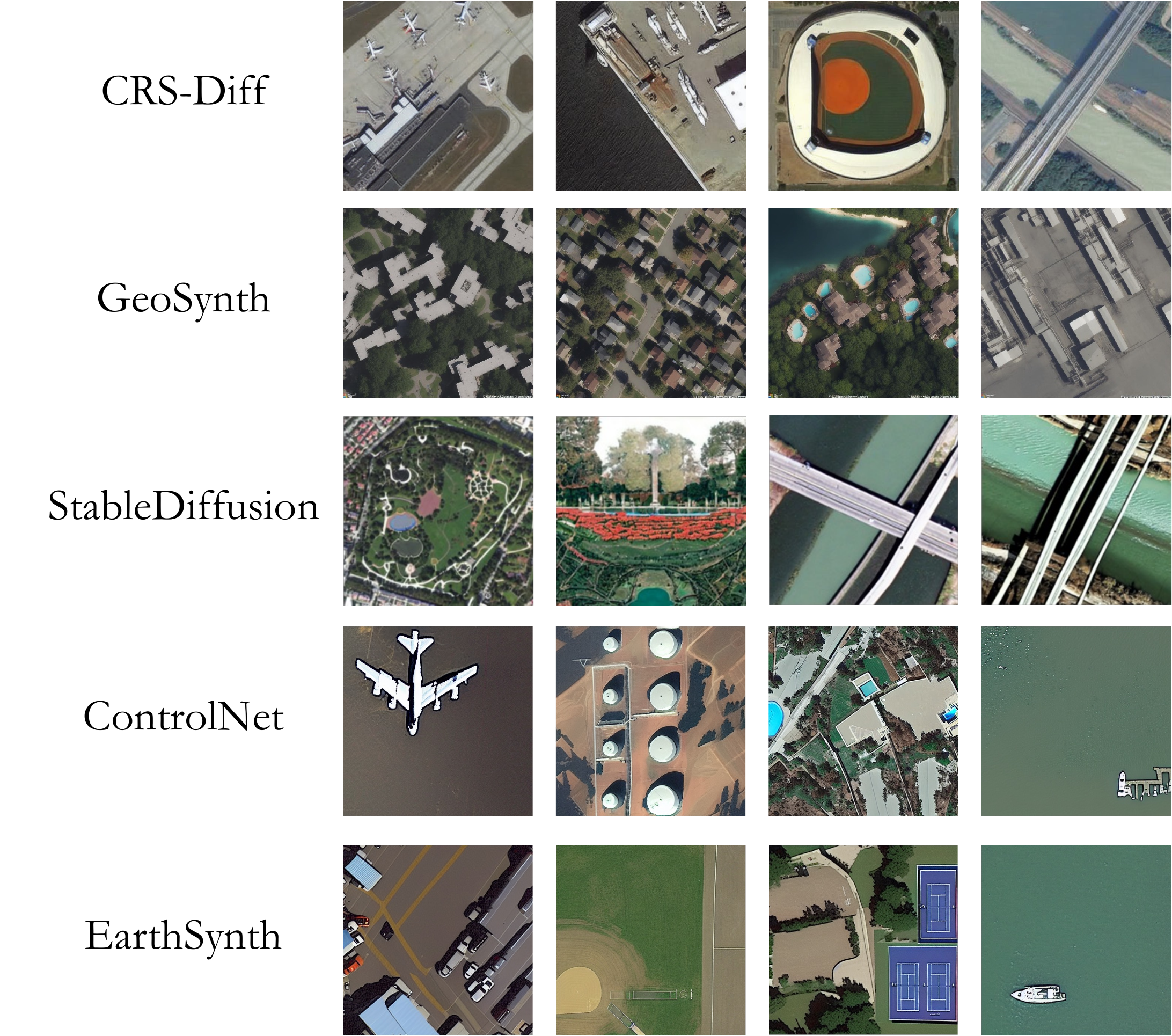}}
% \small
\caption{Comparison of different generation methods.}
\label{ex-fig7}
\end{center}
% \vskip -0.2in
\end{wrapfigure}

Figure~\ref{ex-fig7} compares generated images from remote sensing diffusion models. The focus of the different methods for generating images is described below.

\textbf{CRS-Diff.}
CRS-Diff exhibits relatively limited image quality due to its constrained training data. The restricted diversity and quantity of training samples negatively impact the model’s ability to generalize across various geographic regions and land cover types. As a result, the generated images often lack the visual fidelity and semantic richness observed in outputs from models trained on more comprehensive datasets.

\textbf{GeoSynth.}
GeoSynth is tailored explicitly for generating high-resolution remote sensing images, with training data predominantly sourced from urban areas in the United States. While it excels in producing detailed imagery within this domain, its generative capacity is limited when synthesizing scenes outside this geographic or semantic scope. In particular, its ability to represent diverse land use categories or non-urban environments is relatively weak, restricting its applicability in global or multi-domain remote sensing tasks.

\textbf{Stable Diffusion.}
Although Stable Diffusion demonstrates strong general-purpose image generation capabilities, it lacks mechanisms for spatial layout control. This limitation is critical in remote sensing applications, where the accurate placement and arrangement of objects, such as buildings, roads, and vegetation, are essential for downstream tasks. The inability to control spatial semantics diminishes its utility in structured synthesis scenarios where geospatial coherence and object positioning matter.

\textbf{ControlNet and EarthSynth.}
ControlNet and the proposed EarthSynth model support explicit spatial layout control, allowing for the guided generation of images with well-defined structures and localized semantic targets. This capability is particularly valuable in tasks such as data augmentation, simulation-based training, or synthetic dataset creation for segmentation and detection models. EarthSynth, in particular, further enhances visual realism while preserving layout fidelity, making it a powerful tool for generating structured, high-quality remote sensing imagery across diverse environments and object categories.

\subsubsection{Guidance Scale Analysis}

\begin{figure}[t]
\begin{center}
\centerline{\includegraphics[width=\linewidth]{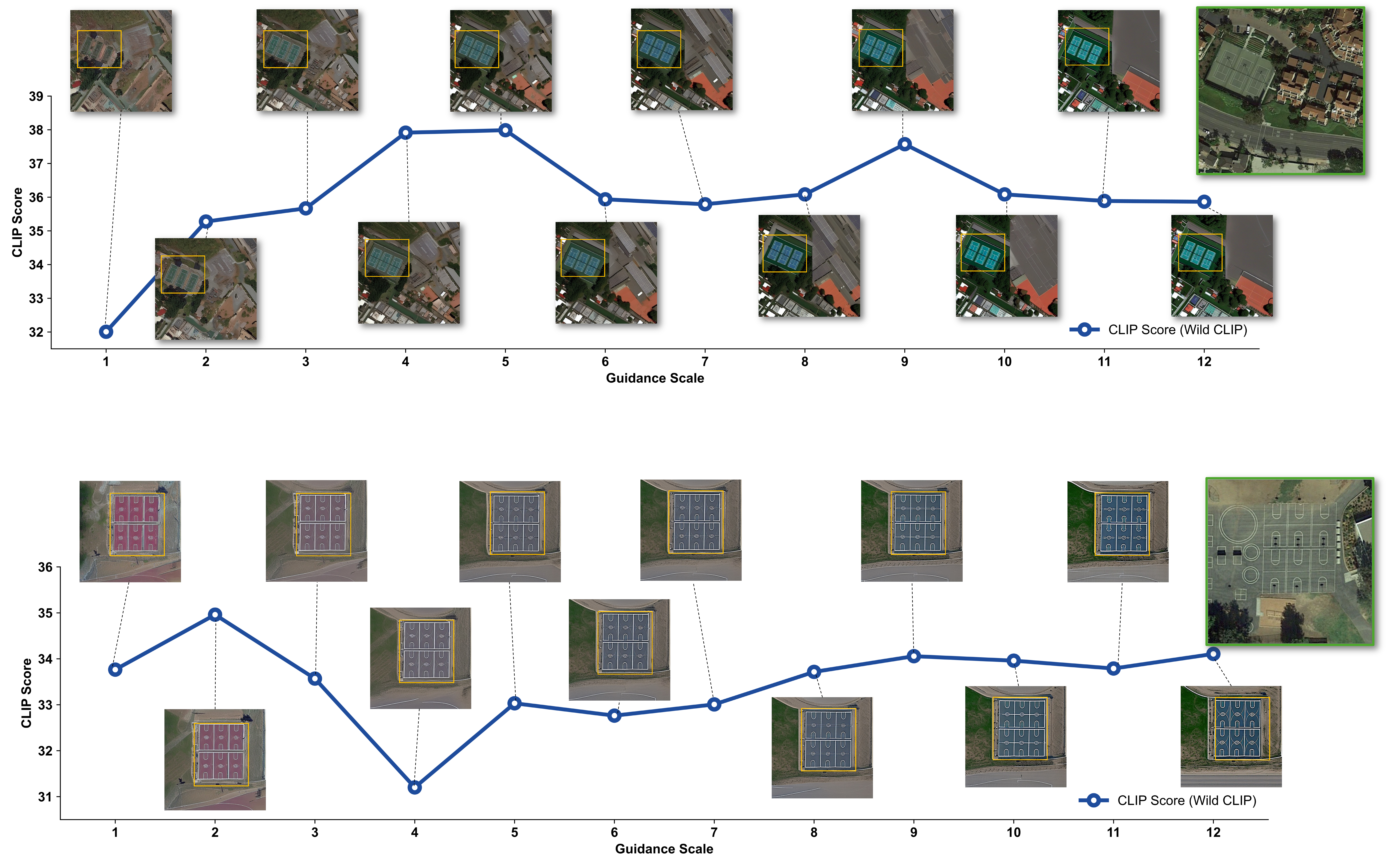}}
\caption{The guidance scale affects the CLIP score across two categories.}
\label{ex-fig5}
\end{center}
\end{figure}

Figure~\ref{ex-fig5} shows how the guidance scale affects the CLIP score. As the scale increases, the CLIP Score rises first, then levels off or slightly decreases. This suggests that moderate guidance scales lead to better alignment between the generated image and the text prompt. In the top examples, representing tennis courts, the CLIP Score reaches its peak around scale 4, with images showing more transparent structure and improved object fidelity. In the bottom examples, representing playgrounds, the score is highest near scale 2 or 3, but they have a poor image generated. Lower scales produce blurry or semantically weak images, while higher scales enhance visual clarity but may reduce diversity. These results indicate that a moderate guidance scale, typically between 3 and 5, balances semantic alignment and image quality well. And we can also find that generation varies across different images, and tuning the guidance scale provides a simple way to control semantic accuracy and visual structure.

\subsection{Limitations} 
We summarize key considerations and limitations of the EarthSynth-180K and EarthSynth as follows:

\textbf{Limited Multispectral Generalization.} This work focuses on generating optical images. However, optical images only cover the visible spectrum and lack the broader spectral information in multispectral images. This limits their use in vegetation monitoring, material classification, and environmental analysis. Although the EarthSynth framework aims for cross-satellite generalization, training and testing are done on the EarthSynth-180K dataset without standard multispectral satellites. Therefore, generalization across sensors with different spectral features is unproven. Extending the framework to handle multispectral data is an important future direction.

\textbf{More Training Cost.} EarthSynth's CF-Comp strategy assumes equal training costs, but training large generative diffusion models is time- and resource-intensive. Since the CF-Comp strategy involves sample combinations in each batch, these setups may not be feasible in resource-limited or costly environments, making some comparisons more theoretical.

\end{document}